\documentclass[lettersize,journal]{IEEEtran}
\usepackage{amsthm}
\usepackage{soul}
\usepackage[normalem]{ulem}
\usepackage[hidelinks]{hyperref}
\usepackage{bookmark}
\usepackage[utf8]{inputenc}
\usepackage[english]{babel}
\usepackage{mathtools}

\usepackage[dvipsnames]{xcolor}
\usepackage{amsfonts,amssymb,amsmath,color}
\usepackage{graphicx}
\usepackage{cite}
\usepackage{placeins}
\usepackage{graphicx}
\usepackage{algorithm}
\usepackage[]{algpseudocode}
\usepackage{todonotes}
\usepackage{balance}
\usepackage{tikz}
\usetikzlibrary{shapes,arrows,calc}
\usepackage{xpatch}

\newcommand\oprocendsymbol{\hbox{$\square$}}
\newcommand\oprocend{\relax\ifmmode\else\unskip\hfill\fi\oprocendsymbol}
\def\eqoprocend{\tag*{$\square$}}

\newtheorem{theorem}{Theorem}[section]

\newtheorem{remark}[theorem]{Remark}

\newtheorem{assumption}[theorem]{Assumption}

\makeatletter
\newcommand{\StatexIndent}[1][3]{%
  \setlength\@tempdima{\algorithmicindent}%
  \Statex\hskip\dimexpr#1\@tempdima\relax}
\makeatother

\renewcommand{\inf}{\operatornamewithlimits{inf\vphantom{p}}}

\renewcommand{\lim}{\operatornamewithlimits{lim\vphantom{p}}}

\newcommand{\lexmin}{\mathop{\rm lex\text{-}min}}

\newcommand{\real}{{\mathbb{R}}}
\newcommand{\binset}{\{0,1\}}
\renewcommand{\natural}{{\mathbb{N}}}
\newcommand{\integer}{{\mathbb{Z}}}
\newcommand{\until}[1]{\{1,\ldots,#1\}} 

\newcommand{\map}[3]{#1: #2 \rightarrow #3}

\newcommand{\Prox}{\texttt{Prox}}

\DeclareMathOperator{\col}{\text{col}}

\newcommand{\argmin}{\mathop{\rm argmin}}

\newcommand{\subj}{\textnormal{subj. to}}
 
\newcommand{\nbrs}{\mathcal{N}}
\newcommand{\conv}[1]{\textnormal{conv}(#1)}

\newcommand{\smallsum}{\textstyle\sum\limits}
\newcommand{\0}{\mathbf{0}}

\newcommand\norm[1]{\left\lVert#1\right\rVert}

\newcommand{\EE}{\mathcal{E}}
\newcommand{\EEt}{\mathcal{E}^t}

\newcommand{\GG}{\mathcal{G}}
\newcommand{\GGt}{\mathcal{G}^t}

\newcommand{\NN}{\mathcal{N}}
\newcommand{\WeightAdj}{\mathcal{A}}

\newcommand{\eqdef}{\coloneqq}

\newcommand{\innbrs}{\mathcal{N}}

\newcommand{\n}{n}
\newcommand{\N}{N}
\newcommand{\dcou}{S}
\newcommand{\set}{\mathbb{I}}

\usepackage{xparse}
\NewDocumentCommand{\xag}{O{}O{}}{x_{#1}^{#2}}
\NewDocumentCommand{\hxag}{O{}O{}}{\hat{x}_{#1}^{#2}}
\NewDocumentCommand{\fag}{O{}O{}}{f_{#1}^{#2}}
\NewDocumentCommand{\nag}{O{}}{\n_{#1}}
\NewDocumentCommand{\X}{O{}}{X_{#1}}
\NewDocumentCommand{\aij}{O{}}{a_{ij}^{#1}}

\newcommand{\constrcoup}[1]{Multi-Robot Constraint-Coupled #1}
\newcommand{\aggropt}[1]{Multi-Robot Aggregative #1}
\newcommand{\xjneigh}{x_{\innbrs_i^t}}
\newcommand{\stepsize}{\gamma}

\def\er/{Erd\H{o}s-R\'enyi}

\graphicspath{{figs/}}

\makeatletter
\newcommand{\specialcell}[1]{\ifmeasuring@#1\else\omit$\displaystyle#1$\ignorespaces\fi}
\makeatother

\newcommand{\dist}{\texttt{dist}}

\newcommand\lexge{\stackrel{\mathclap{\tiny\normalfont\mbox{L}}}{>}}

\begin{document}

\title{\LARGE A Tutorial on Distributed Optimization for Cooperative Robotics:\\ from Setups and Algorithms to Toolboxes and Research Directions}

\author{Andrea Testa, Guido Carnevale and Giuseppe Notarstefano \thanks{
This work was supported by
the European 
  Research Council (ERC) under the European Union's Horizon 2020 research 
  and innovation programme (grant agreement No 638992 - OPT4SMART).
A. Testa, G. Carnevale and G. Notarstefano are with the Department of Electrical,
    Electronic and Information Engineering, University of Bologna, Bologna, Italy.
    \texttt{\{a.testa, guido.carnevale, giuseppe.notarstefano\}@unibo.it}.
  }%
}

\maketitle

\begin{abstract}
Several interesting problems in multi-robot systems can be cast in the framework of distributed optimization. Examples include multi-robot task allocation, vehicle routing, target protection, and surveillance. While the theoretical analysis of distributed optimization algorithms has received significant attention, its application to cooperative robotics has not been investigated in detail. In this paper, we show how notable scenarios in cooperative robotics can be addressed by suitable distributed optimization setups. Specifically, after a brief introduction on the widely investigated consensus optimization (most suited for data analytics) and on the partition-based setup (matching the graph structure in the optimization), we focus on two distributed settings modeling several scenarios in cooperative robotics, i.e., the so-called constraint-coupled and aggregative optimization frameworks. For each one, we consider use-case applications, and we discuss tailored distributed algorithms with their convergence properties. Then, we revise state-of-the-art toolboxes allowing for the implementation of distributed schemes on real networks of robots without central coordinators. For each use case, we discuss its implementation in these toolboxes and provide simulations and real experiments on networks of heterogeneous robots.
\end{abstract}


\section{Introduction}
\label{sec:intro}

\begin{figure}[!ht]
	\centering
	\includegraphics{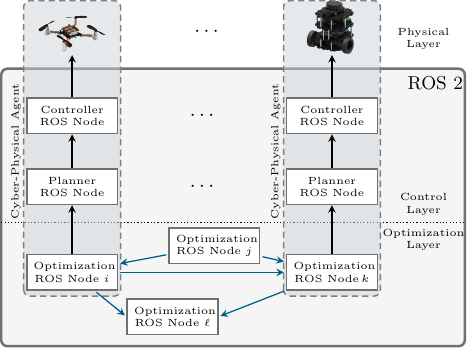}
	\caption{Example of a robotic network. A team of heterogeneous robots cooperatively solves an optimization problem, leveraging the ROS 2 infrastructure. Local planning and actuation follow the result of the optimization.}
	\label{fig:graph}
\end{figure}

In the last decade, the optimization community has developed a novel
theoretical framework to solve optimization problems over networks of
communicating agents.
In this computational setting, agents (e.g., processors or robots) cooperate with the aim of optimizing a common performance index without resorting to a central computing unit.
The key assumption is that each agent is aware of only a small part of the optimization problem, and can communicate only with a few
neighbors~\cite{nedic2018distributed,nedic2018network,nedic2018distributed_b,notarstefano2019distributed,yang2019survey}.
We refer the reader to Figure~\ref{fig:graph} for an illustrative representation.
These features of distributed optimization are attractive for a wide range of application scenarios arising in different scientific communities.
Popular examples can be found, e.g., in the context of cooperative
learning~\cite{park2021communication}, distributed signal
processing~\cite{dimakis2010gossip}, and smart energy systems~\cite{yu2016smart,molzahn2017survey}.
On the other hand, cooperative robotics has become a pervasive technology in a wide range of fields~\cite{dorigo2021swarm}.
Moreover, robotics is increasingly enhanced toward autonomy by relying on typical tools from artificial intelligence~\cite{haidegger2022robot,nanjangud2018robotics}.

The multidisciplinarity and the potential of combining distributed optimization
and multi-robot systems can be harnessed synergistically within the framework of
distributed cooperative robotics. 
Indeed, cooperative multi-robot systems are attracting more and more attention, as they
can be used to perform complex tasks in an efficient manner. 
Despite the increasing popularity of distributed optimization in the context of cooperative learning
and data analytics, this novel computation paradigm has not been extensively
exploited to address cooperative robotics tasks~\cite{rizk2019cooperative}. 
Indeed, several problems of interest in cooperative robotics, such as surveillance, task allocation, and multi-robot (optimal) control can be written as optimization problems in which the goal is to
minimize a certain cost function while satisfying a number of constraints.
In this paper, we show how several problems from cooperative robotics can be solved using distributed optimization techniques.

\subsection{Contributions} The main contribution of this tutorial paper is
twofold. As a first contribution, we show how distributed optimization can be a
valuable modeling and solution framework for several applications in cooperative
robotics. Specifically, after a brief revision of the widely investigated
consensus optimization setup, the paper focuses on two problem classes, namely
constraint-coupled and aggregative optimization, modeling several complex
(control, decision-making, and planning) tasks in which robots do not need to
necessarily achieve a consensus.
For each setting, we show how notable problems arising in multi-robot
applications can be tackled by leveraging these optimization frameworks. As for
the constraint-coupled optimization setup, we provide as main examples the
task-assignment, pickup-and-delivery, and battery-charge scheduling problems. For
the aggregative optimization setting, instead, we consider multi-robot
surveillance and resource allocation as motivating scenarios.
Then, we review different distributed algorithms tailored to these optimization setups and show how they provide an optimal solution in a scalable fashion.
Indeed, an important property of these algorithms is that the local computation
does not depend on the number of robots in the network. Moreover, each robot
retrieves only a useful subset of the entire optimization variable. For these
reasons, they are amenable to onboard implementations in limited memory and
computation settings.
For each one of the presented algorithms, we also report the technical assumptions needed to claim their specific convergence properties.
As a second contribution, we revise toolboxes that allow users to implement
distributed optimization algorithms in multi-robot networks. The toolboxes allow
users to solve distributed optimization settings without central coordination
implementing WiFi communications. For each application scenario, we discuss
implementations of the distributed optimization algorithms via toolboxes based
on the novel Robotic Operating System (ROS) 2 and show simulations and
experiments on real teams of heterogeneous robots.
%
%
While other tutorial and survey papers are available on distributed
  \emph{consensus optimization},~\cite{nedic2018distributed,nedic2018network,nedic2018distributed_b},
  also applied to cooperative
  robotics,~\cite{shorinwa2023distributed_a,shorinwa2023distributed_b,halsted2021survey}, to the
best of the authors' knowledge, this is the first tutorial paper targeting
constraint-coupled and aggregative optimization settings. These frameworks model
a large class of multi-robot decision-making, planning, and control scenarios in
which the multi-robot system goal can be very general and is not limited to
reaching an agreement on a (possibly large) solution vector, as for consensus
optimization.
Moreover, to the best of the authors' knowledge, toolboxes tailored for distributed optimization are not discussed in other related surveys and tutorials.

\subsection{Optimization in Robotic Applications}

In this section, we review well-known robotic applications modeled as
optimization problems.

A widely investigated setup, related to distributed optimization, is the formation
control (or flocking), i.e., the framework in which robots of a team coordinate
their motion to maintain a desired shape, see the
survey~\cite{beaver2021overview}.
In~\cite{hu2020convergent}, a distributed control algorithm embedded with an
assignment switching scheme is proposed.
In~\cite{alonso2017multi}, the authors propose a method for the navigation of a
team of robots while re-configuring their formation to avoid static and dynamic
obstacles.
Assignment and formation placement are simultaneously addressed
in~\cite{mosteo2017optimal}.
Formation control is used in~\cite{tron2016distributed} for application in
vision-based measurements.
A neural-dynamic optimization-based nonlinear Model Predictive Control (MPC) approach
is used in~\cite{xiao2016formation} in the context of formation control.
The work~\cite{notarstefano2011distributed} shows that formation control can be
addressed through an abstract optimization approach.
In~\cite{derenick2007convex}, formation control problems are cast as convex
optimization programs, while~\cite{raffard2004distributed} exploits a dual
decomposition approach.
Some of the above references can be derived as distributed control laws without
optimization schemes. Other works, as,
e.g.,~\cite{mosteo2017optimal,derenick2007convex,raffard2004distributed} use
optimization schemes that exploit the sparsity of the problem.  Indeed, the
formation control setup may be cast as a partition-based optimization problem,
which is discussed in Section~\ref{sec:consensus_opt}.
  %

Another area of interest is multi-robot path planning, where the goal is to
generate trajectories for a team of robots that minimize a performance index.
A decentralized and distributed trajectory planner is proposed
in~\cite{tordesillas2021mader}.
Trajectory planning for multi-quadrotor autonomous navigation is studied
in~\cite{zhang2021decentralized}.
In~\cite{li2020efficient}, trajectory planning is studied for teams of
non-holonomic mobile robots.
The work~\cite{park2020efficient} proposes an algorithm for trajectory planning
combining optimization- and grid-based concepts.
The path-planner method designed in~\cite{das2020multi} combines particle swarm
optimization with evolutionary operators, while~\cite{purcaru2013multi} combines
it with a gravitational search algorithm.
The path-planning problem of a team of mobile robots is studied
in~\cite{kapoutsis2017darp} to cover an area of interest with prior-defined
obstacles.
Multi-robot path planning is addressed in~\cite{wang2015novel} through a
multi-objective artificial bee colony algorithm.
In~\cite{riazi2015energy}, path planning is used to reduce the energy
consumption of an industrial multi-robot system.
In~\cite{bhattacharya2011distributed}, multi-robot path planning is cast as a distributed optimization problem with pairwise constraints.
The work~\cite{luna2010network} deals with path-planning for robots moving on a
road map guided by wireless nodes.
Path-planning problems are often solved by leveraging ad-hoc schemes that do not
involve optimization procedures. However, in order to keep into account
dynamics constraints and interaction among robots, these problems fall in the
class of scheduling problems modeled via the constraint-coupled optimization
framework discussed in Section~\ref{sec:constraint_coupled}.

Another popular area of robotic applications involves cooperative coverage,
where multi-robot systems aim to cover a given area in an optimal manner for,
e.g., mapping, monitoring, or surveillance purposes.
In~\cite{benevento2020multi}, by using tools from Bayesian Optimization with
Gaussian Processes, a team of robots optimizes the coverage of an unknown
density function progressively estimated.
In~\cite{kapoutsis2019distributed}, a distributed algorithm is proposed and then
applied in the context of multi-robot coverage.
The authors of~\cite{notomista2019constraint} investigate the concept of
\emph{survivability} in robotics, with an application in the context of coverage
control.
In~\cite{karapetyan2018multi}, monitoring of aquatic environments using
autonomous surface vehicles is considered.
Distributed asynchronous coordination schemes for multi-robot coverage are
proposed in~\cite{cortes2004coverage}.
Coverage problems are often related to surveillance problems. In this paper, we
show how surveillance problems can be addressed by leveraging the recent
aggregative optimization framework, see Section~\ref{sec:aggregative}.

Other popular multi-robot applications can be found in the context of task allocation, where the aim is to assign a set of tasks to a team of cooperative
robots according to an underlying performance criterion.
In~\cite{camisa2022multi}, the pickup-and-delivery vehicle routing problem is
addressed through a distributed algorithm based on a primal decomposition
approach.  Dynamic vehicle routing problems are solved using centralized
optimization techniques in~\cite{bullo2011dynamic,bopardikar2010dynamic}.
In~\cite{testa2021generalized}, a distributed branch-and-price algorithm is
proposed to solve the generalized assignment problem in a cooperative way.
A time-varying multi-robot task allocation problem is studied
in~\cite{nam2019robots} by using a prior model on how the cost may change.
The article~\cite{bai2019efficient} studies the precedence-constrained task
assignment problem for a team of heterogeneous vehicles.
%
%
In~\cite{testa2019distributed}, mixed-integer linear programs are addressed
through a distributed algorithm based on the local generation of cutting planes,
the solution of local linear program relaxations, and the exchange of active
constraints.
%
Authors in~\cite{smith2009monotonic} propose a distributed algorithm for target
assignment problems in multi-robot systems.
A distributed version of the Hungarian method is proposed
in~\cite{chopra2017distributed} to solve a multi-robot assignment problem.
In~\cite{luo2015distributed}, the authors propose distributed algorithms to
perform multi-robot task assignments where the tasks have to be completed within
given deadlines since each robot has limited battery life.
In~\cite{settimi2013subgradient}, the multi-robot task assignment is addressed
by considering also the local robot dynamics.
A distributed version of the simplex algorithm is proposed
in~\cite{burger2012distributed} for general degenerate linear programs.
In~\cite{karaman2008large}, a team of Unmanned Aerial Vehicles (UAVs) addresses
multi-robot tasks and/or target assignment problems via a dual simplex ascent.
In~\cite{castanon2003distributed}, the case of discovery of new tasks or potential failures of some robots is addressed.
In~\cite{williams2017decentralized}, both topological and abstract constraints
are used for task allocation problems by applying the combinatorial theory of
the so-called matroids. 
Task allocation with energy constraints is studied
in~\cite{hartuv2018scheduling} for a team of drones.
In this tutorial, in Section~\ref{sec:task_allocation}, task allocation problems
are formalized in the form of constraint-coupled optimization programs.
Coherently, Section~\ref{sec:exp_cco1} provides experiments in which task
allocation is addressed through a distributed dual decomposition algorithm.

Finally, increasing attention has been received by MPC approaches for multi-robot systems.
MPC is used in~\cite{torrente2021data} to control a team of quadrotors affected
by aerodynamic effects modeled using Gaussian Processes.
In~\cite{nubert2020safe}, a robust MPC approach is used for a robotic
manipulator and its computational effort is mitigated using a neural network.
In~\cite{leung2006planning,luis2020online}, trajectory generation for multi-robot systems is performed through MPC.
Localization and control problems involving multi-robot systems are concurrently
addressed in~\cite{mehrez2017optimization} using MPC.
In~\cite{van2017distributed}, MPC is used in the context of multi-vehicle
systems moving in formation, while,
in~\cite{nascimento2013multi,nascimento2016multi}, is used for multi-robot
target tracking.
In Section~\ref{sec:pev}, we consider a distributed MPC problem arising in the
context of optimal charging of electric robots and we show that the obtained
problem matches the constraint-coupled formulation.
This formulation allows us to address this setup by leveraging a distributed
algorithm for constraint-coupled optimization, see Section~\ref{sec:exp_cco2}
for the related numerical simulations.

In cooperative robotic surveillance, a team of robots needs to find the optimal
configuration to protect a target from a set of intruders, see the
surveys~\cite{robin2016multi,pasqualetti2012cooperative} for a related overview.
In~\cite{duan2021markov}, the problem is addressed by using tools from Markov
chains.
In~\cite{stump2011multi}, the multi-robot surveillance is cast as a vehicle
routing problem with time windows, i.e., a class of problems widely studied in
operations research.
Multi-robot surveillance is also related to cooperative target tracking, see,
e.g.,~\cite{derenick2009optimal,kamath2007triangulation,tang2005motion}.
In Section~\ref{sec:aggro1}, we model multi-robot surveillance problems by
resorting to the aggregative optimization framework.
Based on this formulation, in Section~\ref{sec:aggro1_exp}, we show related
experiments using a distributed algorithm tailored for the aggregative setup.

\subsection{Toolboxes for Cooperative Robotics}
As for toolboxes to run simulations and experiments on multi-robot teams,
several computational architectures have been proposed, most of them based on
ROS, see, e.g.,~\cite{paxton2017costar}.
%
The works~\cite{grabe2013telekyb,meyer2012comprehensive} propose packages to
simulate and control UAVs. 
Authors in~\cite{dos2016implementing} propose a
software architecture for task allocation scenarios.
%
The Robotarium~\cite{wilson2020robotarium} instead is a proprietary platform
allowing users to test and run control algorithms on robotic teams.  Recently,
ROS~2 is substituting ROS as a new framework for robot programming.  Authors
in~\cite{erHos2019integrated,erHos2019ros2} propose ROS~2 frameworks for
collaborative manipulators, while works in~\cite{mai2022driving,kaiserros2swarm}
propose toolboxes for ground mobile robots executing cooperative tasks.
However, these frameworks are tailored for specific tasks and
platforms. Moreover, cooperation and communication among robots are often
neglected or simulated by computing demanding all-to-all broadcasts.
As for works addressing the development of toolboxes for distributed,
cooperative robotics, authors in~\cite{testa2021choirbot} propose a ROS~2
architecture to run simulations and experiments on multi-robot teams in which
robots are able to communicate with a few neighbors on a WiFi mesh. Leveraging
the DISROPT package~\cite{farina2020disropt}, the architecture in~\cite{testa2021choirbot} allows users to also
model and solve distributed optimization problems. Recently, authors
in~\cite{pichierri2023crazychoir} tailored the work in~\cite{testa2021choirbot}
for fleets of Crazyflie nano-quadrotors.

\subsection{Related Tutorial and Survey Papers}
Several survey and tutorial works have been proposed in the literature
concerning the solution of optimization problems via distributed
algorithms. Tutorial~\cite{notarstefano2019distributed} and survey
~\cite{yang2019survey} provide an introductory overview of foundations in
distributed optimization with some example applications in data analytics and
energy systems. The works
in~\cite{nedic2018distributed,nedic2018network,nedic2018distributed_b} focus on
theoretical methods for consensus optimization in multi-agent systems. The
survey~\cite{scutari2018parallel} is instead centered on big-data optimization,
while the survey~\cite{yang2010distributed} considers also games over
networks. In~\cite{li2022survey}, the main focus is on the online framework.
These surveys take into account different setups arising in distributed
optimization and discuss the role of the communication network in the
convergence of the considered optimization protocols.  However, these works are
not focused on multi-robot applications and do not discuss optimization settings
arising in robotic scenarios such as task allocation, pickup and delivery, and
surveillance.
To the best of the authors' knowledge, few works address distributed
optimization settings in multi-robot scenarios. The
works~\cite{shorinwa2023distributed_a,shorinwa2023distributed_b,halsted2021survey}
consider distributed algorithms to address consensus optimization problems in
which the cost is the sum of local objective functions depending on a common
variable. Finally, a game-theoretic framework for multi-robot systems is
addressed in the survey~\cite{jaleel2020distributed}.

\subsection{Organization}
The paper unfolds as follows. In Section~\ref{sec:preliminaries}, the distributed
optimization model for multi-robot systems is introduced. In
Section~\ref{sec:consensus_opt}, we recall the widely-investigated consensus and
partition-based optimization frameworks. In Section~\ref{sec:constraint_coupled},
we describe the constraint-coupled optimization framework and three applications
in multi-robot settings. In Section~\ref{sec:aggregative}, we describe the
aggregative optimization framework and two applications for multi-robot
networks. In Section~\ref{sec:constr_algo}, we show distributed algorithms to
solve constraint-coupled optimization problems, while distributed schemes for
the aggregative setting are presented in Section~\ref{sec:aggregative_algo}. In
Section~\ref{sec:toolbox}, we review toolboxes to implement distributed
optimization schemes on multi-robot networks and provide simulations and
experiments.  In Section~\ref{sec:research_dir}, we discuss possible future
research directions.

\subsection{Notation}
The set of natural numbers is denoted as $\natural$, while $\real$ denotes the set of real numbers and 
$\real_+$ denotes the set of non-negative real numbers.
We use $\col(v_1, \ldots, v_n)$ to denote the vertical concatenation of the column vectors $v_1, \ldots, v_n$. 
The Kronecker product is denoted by
$\otimes$. 
The identity matrix in $\real^{n\times n}$ is $I_n$.
The column vectors of $N \in \natural$ ones and zeros are denoted by $1_N$ and $0_n$, respectively.
Dimensions are omitted whenever they are clear from the context.
Given a closed and convex set $X \subseteq \real^n$, we use $P_{X}[y]$ to denote
the projection of a vector $y \in \real^n$ on $X$, namely $P_{X}[y] = \argmin_{x
  \in X} \norm{x-y}$, while we use $\dist(y,X)$ to denote its distance from
the set, namely $\dist(y,X) = \min_{x \in X}\norm{x-y}$. 
Given a function of two variables $f: \real^{n_1} \times \real^{n_2} \to \real$,
we denote as $\nabla_1 f \in \real^{n_1}$ the gradient of $f$ with respect to
its first argument and as $\nabla_2 f \in \real^{n_2}$ the gradient of $f$ with
respect to its second argument. 
Given the vector $v \in \real^n$, the symbol $\ln(v) \in \real^n$ denotes the
natural logarithm in a component-wise sense. 
Given $a, b \in \real$ with $a \leq b$, the symbol $[a,b]^n \subset \real^n$
denotes the subset whose belonging vectors $v \triangleq \col(v_1,\dots,v_n) \in
[a,b]^n$ have components $v_i \in [a,b]$ for all $i \in \until{n}$.

\section{Distributed Multi-Robot Setup: from Theoretical Modeling to ROS~2
  Implementation}
\label{sec:preliminaries}
In this section, we first introduce the main theoretical ingredients to model
optimization-based multi-robot frameworks. Then, we show how to convert such a
modeling into a practical implementation by relying on the novel robotic
operating system ROS~2.

\subsection{Theoretical Modeling}
Throughout this tutorial, we consider a set $\set \triangleq \until{N}$ of 
robots endowed with computation and communication capabilities. 
%
Robots aim to self-coordinate by cooperatively solving optimization problems.
More in detail, each robot $i \in \set$ is equipped with a decision variable
$\xag[i] \in \real^{\nag[i]}$ and a related performance index $\fag[i]$ depending on the
local decision variable and, possibly, on the other ones. In the following sections, we provide examples in which the local cost function of the generic robot $i$ may depend on the optimization variables of other robots or on aggregation optimization variables.
The aim of the robotic team is to cooperatively optimize the decision variables
$x_1, \dots, x_N$ with respect to a global performance index $\sum_{i=1}^\N \fag[i]$.
Moreover, since the decision variables $\xag[i]$ are related to the robots' states
and control inputs, then they may be constrained to belong to local sets, i.e.,
it must hold $\xag[i] \in \X[i]$ for some $\X[i] \subseteq \real^{\nag[i]}$. 
Moreover, in some applications, bounded, shared resources are available so that constraints coupling all variables are needed.
%
%
A distinctive feature of \emph{distributed} optimization is that each robot
knows only a part of the optimization problem and can interact only with
\emph{neighboring robots}.
Thereby, in order to solve the problem, robots have to exchange some kind of
information with neighboring robots and perform local computations based on
\emph{local} data.

From a mathematical perspective, this communication can be modeled by means of a
time-varying digraph $\GG^t=(\set,\EE^t)$, where $t\in\natural
$ is a universal slotted time representing temporal information on the graph
evolution, while $\EE^t \subset \set \times
\set$ represents the set of edges of $\GG^t$.
A digraph $\GGt$ models inter-robot communications in the sense that there is an edge
$(i,j) \in \EEt$ if and only if robot $i$ can send information to robot
$j$ at time
$t$.  In this tutorial, we will denote the set of \emph{in-neighbors} of robot
$i$ at time $t$ by $\innbrs_{i}^t$, that is, the set of robots $j$ such that there exists an edge $(j,i) \in
\EEt$ (cf. Figure~\ref{fig:graph}). In the picture, at a certain time $t$,
	robot $i$ is subscribed to a topic from robot $j$, so that $\innbrs_{i}^t=\{j\}$.
	Meanwhile, it publishes data on a topic read by robots $k$ and $\ell$.
%
%
We now recall basic properties from graph theory which will be 
used in the rest of the tutorial.
If $\EEt$ does not depend on the time $t$, then the graph is called \emph{static}, 
otherwise, it is called \emph{time-varying}. A graph is said to be \emph{undirected} if 
for each edge $(i,j)\in\EEt$ it stands $(j,i)\in\EEt$ for all $t \in \natural$. Otherwise,
the graph is said to be \emph{directed} and is also called digraph.
A static digraph is said to be \emph{strongly connected} if there exists a directed
path connecting each pair $(i,j) \in \set \times \set$.
A time-varying digraph $\GGt$ is said to be
\emph{jointly strongly connected} if, for all
$t \in \natural$, the graph constructed as $(\set,\bigcup_{\tau=t}^{\infty} \EE^\tau)$ is strongly connected.
Finally, $\GGt$ is said to be
\emph{$L$-strongly connected} if there exists an integer 
$L \geq 1$ such that, for all $t \in \natural$, 
the graph defined as $(\set,\bigcup_{\tau=t}^{t+L-1} \EE^\tau)$ is
strongly connected.
Given a digraph $\GGt$, it is possible to associate to it, at each $t$, a so-called weighted adjacency matrix $\WeightAdj^t\in\real^{\N\times\N}$. We denote by $\aij[t]$ the $(i,j)$-th entry of $\WeightAdj^t$ at time $t$. It holds $\aij[t]>0$ if $(i,j)\in\EEt$, and $\aij[t]=0$ otherwise. A matrix $\WeightAdj \in \real^{N \times N}$ is said to be doubly stochastic if it holds $\WeightAdj1_N = 1_N$ and $1_N^\top\WeightAdj = 1_N^\top$.

\subsection{Hardware-Software Implementation: a ROS-2 perspective}
In real experiments, robots are controlled by a set of inter-communicating
processes. These processes may read data from sensors, implement decision,
planning, and control schemes, and may exchange data with other robots or with
the environment by means of, e.g., WiFi communication.
Following this line, in this tutorial, we consider each robot as a cyber-physical system in which
the aforementioned actions are performed by a set of ROS~2 independent processes (also called \emph{nodes}).
In Figure~\ref{fig:graph}, we depict a network of robots, each one with its set of nodes. Specifically, a Controller Node handles the implementation of control schemes for the robot. The Planner Node implements trajectory planning schemes, while the Optimization Node implements the distributed optimization schemes needed to solve a complex problem.
In ROS~2, nodes can be deployed on independent computing units and can exchange data according to different protocols. 
We focus on the \emph{publish-subscribe} protocol in which nodes can act in two ways, i.e., publisher
and subscriber.
When a node acts as a publisher, it sends data over a dedicated \emph{topic} on the TCP/IP stack. Nodes interested in the
data can then subscribe to the topic to receive it as soon as it is published. 
Nodes can act simultaneously as publishers and subscribers
on different topics (cf. Figure~\ref{fig:graph}). Moreover, they can change almost at run-time the topics they are subscribed to.
Interestingly, ROS~2 implements a broker-less publish-subscribe protocol in which nodes are able
to find and exchange data without the need for a central coordinating unit.
We refer the reader to Figure~\ref{fig:graph} for an illustrative example.

\section{Consensus Optimization \\and Partition-Based Optimization}
\label{sec:consensus_opt}
In this section, we briefly recall the consensus optimization and
partition-based optimization frameworks. As we detail next, these two frameworks
have been extensively studied from a theoretical perspective. Thus, in this
paper, we briefly recall them and, instead, we focus on two other setups that
model a larger set of multi-robot scenarios.
\subsection{Consensus Optimization}
In the so-called \emph{consensus} optimization setup, the objective is to
minimize the sum of $\N$ local functions $\fag[i]:\real^\n\to\real$, each one
depending on a global variable $\xag\in \X\subseteq\real^\n$, namely
\begin{align} 
  \begin{split}
    \min_{\xag} \: & \: \sum_{i=1}^N \fag[i](\xag)
    \\
    \subj \: & \: \xag\in \X.
  \end{split}\label{eq:cost_coupled}
\end{align}
In more general versions of this setup, the set $\X$ can be defined as the intersection of local constraint sets $\X[i]$.
In this setup, robots usually update local solution estimates $\xag[i][t]\in\real^\n$
that eventually converge to a common optimal solution $\xag[][\star]\in\real^\n$
to~\eqref{eq:cost_coupled}.
%
This setup is particularly suited for decision systems performing data analytics
in which data are private and the consensus has to be reached on a common set of
parameters characterizing a global classifier or estimator (e.g., neural
network) based on all data.
This setup has been investigated in several papers for decision-making in multi-agent systems, see,
e.g.,~\cite{notarstefano2019distributed,shorinwa2023distributed_a,shorinwa2023distributed_b,jaleel2020distributed,nedic2018distributed,nedic2018network,li2022survey,nedic2018distributed_b,scutari2018parallel,halsted2021survey,yang2010distributed}
for detailed dissertations.
In cooperative robotics, this scenario can be interesting to reach a consensus on a common quantity, e.g., target estimation.
We refer the reader to the survey~\cite{halsted2021survey} that is centered around this optimization setting.
  %
%
%
In many other tasks within cooperative robotics, we will show the advantages of adopting constraint-coupled and aggregative formulations.
Although consensus optimization provides the most general problem formulation, algorithms designed for such a framework require storing, sharing, and reaching consensus on a decision variable that includes the stack of all robots' estimates.
In contrast, we will show that algorithms tailored for constraint-coupled and aggregative scenarios are particularly appealing because (i) each robot only needs to store a solution estimate and instrumental quantities that are independent of the total number of robots, (ii) consensus is required solely on low-dimensional instrumental quantities (such as the global constraint function or the aggregative variable), and (iii) throughout all iterations, the robots remain unaware of each other's solution estimates.
These features enable the development of distributed algorithms specifically designed for constraint-coupled and aggregative setups, which are particularly advantageous from the point of view of computational, communication, and memory burden that are critical aspects, especially in robotic applications.

\subsection{Partition-Based Optimization}
This framework models problems
  with a \emph{partitioned structure}.  More in detail, let the decision vector
  $\xag\in\real^n$ be in the form
  $\xag = [\xag[1][\top],\ldots,\xag[\N][\top]]^\top$.  Here,
  $\xag[i]\in\real^{\nag[i]}$ and $\sum_{i\in\set}\nag[i]=n$. The subset
  $\xag[i]$ represents information related to robot $i$.  Assume that robots
  communicate according to a static connected, undirected graph. The
  partition-based optimization framework assumes that the objective functions
  and the constraints have the same sparsity structure as the communication
  graph. That is, the function $f_i$ and the constraint set $\X[i]$ depend only
  on $\xag[i]$ and on $\xag[j]$ for all $j\in\innbrs_i$. Thereby,
  partition-based optimization problems are in the form
  \begin{align} 
    \begin{split}
      \min_{\xag[1],\ldots, \xag[\N]} \: & \: \sum_{i=1}^N f_i(\xag[i],\{\xag[j]\}_{j\in\innbrs_i})
      \\
      \subj \: & \: (\xag[i],\{\xag[j]\}_{j\in\innbrs_i})\in \X[i],\quad \forall i\in\set.
    \end{split}\label{eq:partition}
  \end{align}
  This framework has been extensively analyzed from a theoretical perspective,
  see,
  e.g.,~\cite{todescato2020partition,notarnicola2017distributed,notarnicola2016randomized,bastianello2018partition}.
  However, this setup is strictly related to application scenarios in which the
  sparsity of the communication can match the one of functions and
  constraints. In this tutorial, instead, we consider more general application
  scenarios in which the formulation is independent of the network
  structure. Indeed, in multi-robot applications, the communication topology may
  be not related to the optimization sparsity, and it may vary during time due to
  robot movements.

\section{Constraint-Coupled Optimization Setup\\ for Cooperative Robotics}
\label{sec:constraint_coupled}

\begin{figure}[!ht]
  \centering
  \includegraphics{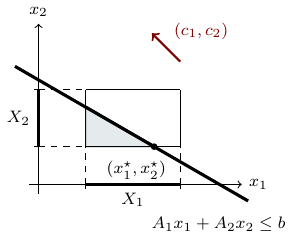}
  \caption{Example of constraint-coupled setup. The two local decision variables
    $\xag[1]\in \X[1]$ and $\xag[2]\in \X[2]$ are coupled by a linear coupling
    constraint $A_1 \xag[1] + A_2 \xag[2]\leq b$. The gray area represents the
    feasible set. The tuple $(c_1,c_2)$ represents the components of the linear cost. The cost direction is represented as a red arrow.}
  \label{fig:constr_coupled}
\end{figure}

In this distributed optimization framework, robots aim at minimizing the sum of
local cost functions $\fag[i]: \real^{\nag[i]} \to \real$, each one depending only
on a local decision vector $\xag[i]\in \X[i]\subseteq \real^{\nag[i]}$.  The
decision vectors of the robots are coupled by means of $S \in \natural$ separable coupling
constraints, each defined by a mapping $g_i:\real^{\nag[i]}\to\real^\dcou$.  Distributed
constraint-coupled optimization problems can be formalized as
\begin{align} 
  \begin{split}\label{eq:constr_coupled}
    \min_{\xag[1],\ldots, \xag[\N]} \: & \: \sum_{i=1}^\N \fag[i](\xag[i])
    \\
    \subj \: & \: \sum_{i=1}^\N g_i(\xag[i]) \leq 0\\
    \: & \: \xag[i]\in \X[i], \forall i \in \set.
  \end{split}
\end{align}
A graphical example of a constraint-coupled linear program with $\N=2$ and $\nag[i]=1$ for
each $i$ is provided in Figure~\ref{fig:constr_coupled}.

In this setup, the size $\n=\sum_{i=1}^\N \nag[i]$ of the entire
decision vector $\xag = \col(\xag[1],\ldots,\xag[\N])\in\real^\n$ usually
increases with the number $\N$ of robots in the network. Thus, scalable
distributed optimization algorithms are designed so that the generic robot $i$
only constructs its local, small portion $\xag[i][\star] \in \real^{\nag[i]}$ of the optimal
decision vector $\xag[][\star] = \col(\xag[1][\star],\ldots,\xag[\N][\star]) \in \real^{\n}$.

\subsection{\constrcoup{1}: Multi-Robot Task Allocation}
\label{sec:task_allocation}
\begin{figure}[!ht]
  \centering
  \includegraphics{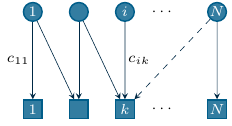}
  \caption{Task assignment problem. Robots are represented by circles, while
    tasks are represented by squares. An arrow from $i$ to $k$ means that robot
    $i$ can perform task $k$, incurring a cost $c_{ik}$.} 
  \label{fig:bipartite}
\end{figure}

In this section, we consider a scenario in which a team of robots has to
negotiate how to optimally serve a set of tasks, see,
e.g.,~\cite{burger2012distributed} for a detailed discussion. This task
allocation problem is a notable example of constraint-coupled
optimization. Here, we focus on a \emph{linear assignment problem}. Consider a
network of $\N$ robots, that has to serve a set of $\N$ tasks. The generic robot
$i$ incurs a cost $c_{ik}$ if it serves task $k$. Moreover, each robot may be
able to serve only a subset of the $\N$ tasks. A graphical representation is
given in Figure~\ref{fig:bipartite}.

The objective is to find a robot-to-task assignment such that all the tasks are
served, each robot serves one task, and the overall cost is minimized.  Let $E_A$
be the set of tuples defined as
$E_A\triangleq\{(i,k) \mid \text{robot } i\text{ can serve task } k\}$. To each robot it
is associated a binary decision vector $\xag[i]\in\real^{\nag[i]}$ with
$\nag[i]=|\{k\mid (i,k)\in E_A\}|$. The $k$-th entry of $\xag[i]$ is equal to
$1$ if the robot serves task $k$ and $0$ otherwise. Let $c_i\in\real^{\nag[i]}$
be the vector stacking the costs $c_{ik}$. The constraint that each robot has to
serve one task can be enforced by the feasible set $\X[i]\triangleq\{\xag[i]\in\{0,1\}^{n_i}\mid 1_{\nag[i]}^\top \xag[i] =
1\}$. Let $H_i\in\real^{\N\times \nag[i]}$ be a matrix obtained by extracting
from a $\N\times \N$ identity matrix all the columns $k$ such that
$(i,k)\in E_A$.
Then, the linear assignment problem can be written as an integer linear program in the form
\begin{align}\label{eq:task_assignment}
  \begin{split}
    \min_{\xag[i],\ldots \xag[\N]} \: & \: \sum_{i=1}^N c_i^\top \xag[i]  \\
    \subj \: & \: \sum_{i=1}^\N H_i \xag[i] = 1_N
    \\
    \: & \: \xag[i]\in \X[i], \forall i \in \set.
  \end{split}
\end{align}

It is worth noting that the linear assignment is a special case of the constraint-coupled optimization problem in which the coupling constraint is a
linear, equality constraint. It is immediate to see that, considering
$g_i(\xag[i])=H_i \xag[i] - 1_N/\N$ and $\fag[i](\xag[i])=c_i^\top \xag[i]$ it is
possible to recover the structure of problem~\eqref{eq:constr_coupled}.
Moreover, exploiting the so-called \emph{unimodularity} of the constraint
matrices, the local sets $\X[i]$ can be replaced with linear, convex sets in the
form
$\X[i]\triangleq\{\xag[i]\in\real^{\nag[i]}\mid 0\leq \xag[i] \leq 1, 1_{\nag[i]}^\top
\xag[i] = 1\}$. In this way, the task assignment can be solved in a distributed fashion as a linear program.

As a motivating example, we consider a distributed service-matching covering scenario as in~\cite{chung2022distributed}. 
More in detail, we leverage the distributed task allocation framework to deploy a team of robots that have to provide a service for a set of targets. 
Targets are distributed according to an unknown distribution $q(x)\in\real$. 
Each robot provides a service (e.g., event detection or power supply) with a Quality of Service (QoS) modeled as a spatial Gaussian distribution. 
In particular, let $p_i\in\real^2$ be the position of the robot in the $2D$ plane and let $\theta_i \in \real$ be its orientation.
Then, the QoS of each robot $i$ is modeled as 
\begin{align*}
  Q_i(x \mid x_i, \theta_i) = \omega_i \NN(x \mid x_i, \Sigma(\theta_i)),
\end{align*}
where $\NN(x \mid \mu_i, \Sigma_i)$ is a Gaussian distribution centered at $\mu_i \in \real^{2}$ with covariance matrix $\Sigma_i(\theta_i) \in \real^{2 \times 2}$, while $\omega_i > 0$ is a scale constant.
Robots have to be deployed so that their collective QoS matches the density distribution $q(x)$ of the targets according to the Kullback-Leibler Divergence (KLD) measure.
In~\cite{chung2022distributed}, authors propose a Distributed Expectation-Maximization algorithm to allow robots to estimate 
the target density distribution $\hat{q}(x)$ using a Gaussian Mixture Model (GMM) in the form
\begin{align*} 
  \hat{q}(x) = \sum_{j=1}^{\N } \pi_j N(x \mid \mu_j, \Sigma_j),
\end{align*}
where $\pi_k$, $\mu_k$, and $\Sigma_k$ are the weight, mean, and covariance matrix of the $k$-th Gaussian component, respectively.
Authors consider a distributed setting in which each robot has access to some samplings of the distribution $p(x)$. 
Running the proposed algorithm, each robot $i$ can estimate a distribution 
\begin{align*}
  \hat{q}_i(x) = \sum_{j=1}^{\N} \pi_{i,j} \NN(x \mid \mu_{i,j}, \Sigma_{i,j}).
\end{align*} 
These distributions may slightly differ from each other due to some early-termination steps in the proposed protocol. 
We refer the reader to~\cite{chung2022distributed} for a detailed discussion. 
The GMM model clusters the targets into a set of subgroups, each one represented by a Gaussian
basis of the GMM.
After estimating the target density, the next step is to assign each robot to a target cluster by minimizing the KLD between the robot QoS and the target cluster distribution. 
For any robot $i$, and for any cluster $j$, let
$c_{ij}(x_i, \theta_i) = D_{KL}( \pi_{i,j} \NN(x|\mu_{i,j}, \Sigma_{i,j}) \| \omega_i \NN(x| x_i, \Sigma(\theta_i)))$, where $D_{KL}(\cdot \| \cdot)$ is the KLD. 
In~\cite{chung2022distributed}, authors propose a closed-form solution to this expression. In particular, the term $c_{ij}$ relates the $i$-th robot QoS to the $j$-th cluster in terms of the KLD metric. 
With these terms at hand, authors consider a task allocation problem in the form of~\eqref{eq:task_assignment}. 
By solving such a problem, it is possible to associate a robot to each cluster. 
By moving the robot into the cluster mean $\mu_j$ and orientating it to the principal axis of $\Sigma_j$, the deployment problem is solved. 
We refer the reader to~\cite{chung2022distributed} for a more detailed explanation and for illustrative examples.


%

\subsection{\constrcoup{2}: Planning of Battery Charging for Electric Robots}
\label{sec:pev}
%
Inspired by the application proposed in~\cite{vujanic2016decomposition}, we
consider a charging scheduling problem in a fleet of $\N$ battery-operated
robots drawing power from a common infrastructure.  This schedule has to satisfy
local requirements for each robot, e.g., the final State of Charge (SoC). The
schedule also has to satisfy power constraints, e.g., limits on the maximum
power flow.
%
%
We assume that charging can be interrupted and resumed.
%
The overall charging period $T$ is discretized into $d$ time steps. For each
robot $i \in \set$, let $u_i\in[0,1]^d$ be a set of decision variables handling
the charging of the robot. That is robot $i$ charges at time step $k$ with a
certain charge rate between $0$ (no charging) and $1$.
%
The $i$-th battery charge level during time is denoted by $e_i\in\real^d$. The
$i$-th battery initial SoC is $E_i^\text{init}$ and, at the end of the charging
period, its SoC must be at least $E_i^\text{ref}$.  

We denote by $E_i^\text{min}$ and $E_i^\text{max}$ the battery's capacity limits, by $P_i$ the maximum charging power that can be fed to the $i$-th robot,
and by $P^{\text{max}}$ the maximum power flow that robots can draw from the
infrastructure.
%
%
Let $C_u^k\in\real$ be the price for electricity consumption at time slot $k$.
Let $\mathbb{T}$ denote the set $\{0,\dots,T-1\}$.  The objective is to minimize
the cost consumption. Then, the optimization problem can be cast as
%
%
\begin{subequations}
  \label{eq:EV_primal_problem}
  \begin{align}
    \underset{\{e_i,u_i\}_{i=1}^\N}{\text{min}} & \sum\limits_{i=1}^\N \sum_{k=0}^{T-1} P_i C_u^k  u_i^k  \\
    \text{subj. to }
    & \sum\limits_{i=1}^\N P_iu_i^k \leq P^{\text{max}} & \forall k\in\mathbb{T} &\label{eq:PEV_coupling}\\
    & e_i^0 = E_i^\text{init} & \forall i\in\mathbb{I}\label{eq:model_e_init}\\
	& e_i^{k+1}= e_i^k +\!P_i \Delta T u_i^k &  \forall i\in\set, k\in\mathbb{T} \label{eq:EV_state_update}\\
    & e_i^T \geq E_i^\text{ref} & \forall i\in\set \label{eq:EV_energ_requirement}\\
    & E_i^\text{min}1_d \leq e_i \leq E_i^\text{max}1_d & \forall i\in\set\\
   & u_i \in [0,1]^d & \forall i\in\set.\label{eq:pev_binary}
  \end{align} 
\end{subequations}
In order to cast~\eqref{eq:EV_primal_problem} as a constraint-coupled
optimization problem, let us define the following quantities.  For each robot
$i$, let $\xag[i]\in\real^{\nag[i]}$ be the stack of $e_i^k,u_i^k$. Also, let
$C_u$ be the stack of $C_u^k$.  Then,
\begin{align}
  \fag[i](\xag[i]) \triangleq c_i^\top\xag[i],
\end{align}
where $c_i$ is a suitable vector in the form $\col(0, P_iC_u)$.
As for the local constraints, let us define for all $i$ the set
\begin{align}
  \X[i] \triangleq \Big\{
  \xag[i] \in\real^{\nag[i]} \mid \text{
  \eqref{eq:model_e_init}--\eqref{eq:pev_binary} are satisfied}
  \Big\}.
\end{align}
%
Finally, all the local variables are coupled by the constraint~\eqref{eq:PEV_coupling}. We underline that, differently from~\eqref{eq:task_assignment}, the problem in~\eqref{eq:EV_primal_problem} does not benefit from the unimodularity property and cannot be solved as a linear problem.

\subsection{\constrcoup{3}: Multi-Robot Pickup-and-Delivery}
\label{sec:pdvrp}
\begin{figure}[htbp]\centering
  \includegraphics[scale=.92]{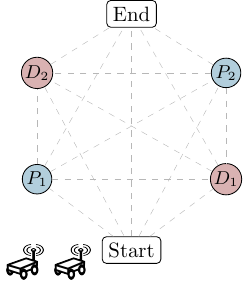}\hfill
  \includegraphics[scale=.92]{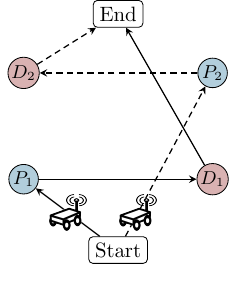}
  \caption{Example PDVRP scenario with two pickup and two delivery requests. Robots are initially located at the ``start'' and must
    end at the terminal node. On the left, dashed, grey lines denote all the possible paths robot can travel. Right: black lines denote the optimal paths for the two robots.}
  \label{fig:pdvrp}
\end{figure}

Inspired by the work~\cite{camisa2022multi}, we consider a network of $\N$ robots that have to serve a number of pickup and delivery requests.  That is, robots
have to pick up goods at some locations and deliver them to other locations.  We
denote by $P \eqdef \until{|P|}$ the index set of pickup requests, while
delivery requests are denoted as $D \eqdef \{|P|+1, \ldots, 2|P|\}$. A good
picked up at location $j\in P$ must be delivered at $j+|P|\in D$. This enforces that
the two requests have to be served by the same robot.  Let $R \eqdef P \cup D$
be the set of all the requests. The objective is to serve all the requests by
minimizing the total traveled distance, see Figure~\ref{fig:pdvrp} for an
example.
%

Each request $j\in R$ is characterized by a service time $d_j\geq 0$, which is the time needed to perform the pickup or delivery operation.
Within each request, it is also associated a load $q_j\in\real$, which is positive
if $j\in P$ and negative if $j\in D$.
Each robot has a maximum load capacity $C_i\geq0$ of goods that can be simultaneously held.
The travel time needed for the $i$-th robot to move from a location $j \in R$ to another
location $k \in R$ is denoted by $t_i^{jk}\geq0$.
In order to travel from two locations $j,k$, the $i$-th robot incurs a cost $c_i^{jk}\in\real_{\geq0}$.
Finally, two additional locations $s$ and $\sigma$ are considered. The first one represents
the mission starting point, while the second one is a virtual ending point.
For this reason, the corresponding demands $q^s, q^\sigma$ and service times
$d^s, d^\sigma$ are set to $0$.
The goal is to construct minimum-cost paths satisfying all transportation requests.
To this end, a graph of all possible paths through the transportation requests
can be defined. We underline that this graph models the optimization problem and
it is not related to the graph modeling inter-robot communication.
  Let $\GG_A=(V_A,\EE_A)$, be the graph with vertex
set $V_A=\{s,\sigma\}\cup R$ and edge set
$\EE_A=\{(j,k)\mid j,k\in V_A, j\neq k \text{ and } j\neq \sigma, k\neq s \}$.
$\EE_A$ contains edges starting from $s$ or from
locations in $R$ and ending in $\sigma$ or other locations in $R$
(cf. Figure~\ref{fig:pdvrp}).
For all edges $(j,k)\in\EE_A$, let $x_i^{jk}$ be a binary variable denoting whether
robot $i\in\set$ is traveling ($x_i^{jk}=1$) or not ($x_i^{jk}=0$) from a location $j$
to a location $k$.
Also, let $B_i^j\in\real_{\geq 0}$ be the optimization variable modeling the time at which
robot $i$ begins its service at location $j$. Similarly, let $Q_i^j\in\real_{\geq 0}$ be the
load of robot $i$ when leaving location $j$.
The PDVRP can be formulated as the following optimization problem
\begin{subequations}%
  \label{eq:PDVRP}%
  \begin{align}
    \min_{
    x, B, Q
    } \: & \: \sum_{i=1}^N \sum_{(j,k) \in \EE_A} c_i^{jk} x_i^{jk}  \label{eq:PDVRP_cost}
    \\
    \subj \:
         & \: \sum_{i=1}^N \sum_{k:(j,k) \in \EE_A} x_i^{jk} \geq 1
           \hspace{2.2cm}  \forall j \in R
           \label{eq:PDVRP_coupling_constr}
    \\
         & \: \text{(flow and precedence constraints)} \hspace{.7cm} \forall i\in\set \label{eq:PDVRP_flow}
    \\
         & \: \text{(temporal and capacity constraints)} \hspace{.5cm} \forall i\in\set \label{eq:PDVRP_cap}\\
         & \: x_i^{jk} \in \binset, B_i,Q_i \in\real
           \hspace{.3cm} \forall i\in\set, (j,k) \in \EE_A.
           \label{eq:PDVRP_local_con8}
  \end{align}
\end{subequations}
For the sake of presentation, we do not report the explicit form of all the
constraints. We refer the reader to~\cite{parragh2008survey} for the detailed
formulation.
%
The PDVRP problem can be cast as a constraint-coupled problem~\eqref{eq:constr_coupled} as follows.
For each $i$, let $x_{i}^{\EE_A}$ be the stack of all the $x_i^{jk}$ for all $(i,j)\in\EE_A$. Similarly,
let $B_i (Q_i)$ be the stack of all the $B_i^j (Q_i^j)$ for all $j\in R$. Finally, let $\xag[i]\in\real^{\nag[i]}$ be the
stack of $x_{i}^{\EE_A}, B_i, Q_i$. With this notation at hand, it stands
\begin{align}
  \fag[i](\xag[i]) \triangleq \sum_{(j,k) \in \EE_A} c_i^{jk} x_i^{jk}.
\end{align}
Note that the constraints~\eqref{eq:PDVRP_flow}--\eqref{eq:PDVRP_local_con8}
are repeated for each index $i$.
Thus, let us define for all $i$ the set
\begin{align}
  \X[i] \triangleq \Big\{
  \xag[i] \in\real^{\nag[i]} \mid \text{
  \eqref{eq:PDVRP_flow}--\eqref{eq:PDVRP_local_con8} are satisfied}
  \Big\}.
\end{align}
%
Finally, notice that~\eqref{eq:PDVRP_coupling_constr} is a constraint coupling the variables $\xag[i]$
of all the robots. Thereby, we model them by setting
\begin{align}
  g_i(\xag[i]) \triangleq \frac{1}{\N} - \sum_{k:(j,k) \in \EE_A} x_i^{jk}.
\end{align}
%

\section{Aggregative Optimization Setup for Cooperative Robotics}
\label{sec:aggregative}

The aggregative optimization setup has been originally introduced in the pioneering works~\cite{li2021distributed_a,li2021distributed_b} in which the first distributed algorithms have been provided.
Such a framework suitably models applications in which the robots take
decisions by simultaneously considering their \emph{local} information and an
\emph{aggregative} description of the network.
Formally, in this setup, the robots are interested in minimizing the sum of
local cost functions $\fag[i]$ but, differently from the constraint-coupled
framework (cf.~Section~\ref{sec:constraint_coupled}), each local function $\fag[i]$
depends both on (i) the associated decision variable $\xag[i]$ and (ii) an
aggregation $\sigma(\xag) = \sigma(\col(\xag[1],\ldots \xag[\N]))$ of all the
decision variables of the network, for a suitable function
$\sigma: \real^{\nag} \to \real^{d}$.
That is, the aggregative optimization problem is defined as
\begin{align}
  \begin{split}
    \min_{\xag[1],\ldots \xag[\N]} \: & \: \sum_{i =1}^\N \fag[i](\xag[i],\sigma(\xag)),
    \\
    \subj \: & \: \xag[i] \in \X[i], \forall i \in \set,
  \end{split}
	\label{eq:aggregative_problem}
\end{align}
where, for all $i \in \set$, $\fag[i] : \real^{\nag[i]} \times \real^d \to \real$ is
the objective function of robot $i$, $\X[i] \subseteq \real^{\nag[i]}$ its
feasible set, and, given $\nag \triangleq \sum_{i=1}^\N \nag[i]$, the aggregative
variable $\sigma(x)$ reads as
\begin{align*}
  \sigma(x) \triangleq \frac{\sum_{i=1}^\N \phi_i(\xag[i])}{\N},
\end{align*}
where $\phi_i : \real^{\nag[i]} \to \real^d$ is the $i$-th contribution to the
aggregative variable.
According to the distributed paradigm, we assume each robot is aware only of
$\fag[i], \phi_i$, and $\X[i]$, thus not having knowledge of other robot data.
%

\subsection{\aggropt{1}: Target Surveillance}
\label{sec:aggro1}
\begin{figure}[!ht]
  \centering
  \includegraphics{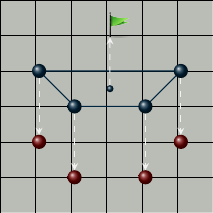}
  \caption{Example of the target surveillance problem. Robots are represented as blue big circles, while the target is represented as a flag. The small blue circle represents the robots' barycenter. Red circles represent adversaries. Each robot wants to move towards a certain adversary, while the barycenter is steered near the target (white arrows).}
  \label{fig:aggregative_example}
\end{figure}

\begin{figure}[H]
  \centering
  \includegraphics[width=.7\columnwidth]{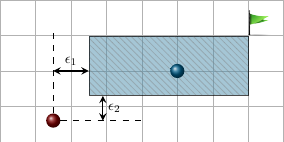}
  \caption{Example of the constraint set for one of the robots $i\in
    \set$. Robot $i$ (blue sphere), can move in the constraint set as
    in~\eqref{eq:agg_con1} and~\eqref{eq:agg_con2} between the adversary (red
    sphere) and the target (green flag).} \label{fig:constr}
\end{figure}

We now introduce a motivating example, namely the \emph{Target Surveillance}
problem, that can be cast into a distributed aggregative optimization
problem. An illustrative example is given in
Figure~\ref{fig:aggregative_example}.
%
In this scenario, a team of $\N$ robots has to protect a target from a set of
$\N$ intruders.
We denote by $\xag[i]\in\real^3$ the position of robot $i\in\set$ and by
$r_0 \in\real^3$ the position of the target.
Each robot wants to stay as close as possible to its designated intruder, but
simultaneously not to drift apart from the target so as not to leave it
unprotected.
Moreover, we impose that each defending robot $i \in \set$ poses itself between
the corresponding intruder and the target to protect.
A 2D example of this constraint set is shown in Figure~\ref{fig:constr}.
%
%
In this setting, we assume that each robot $i$ only knows its position and the
position $r_i \in\real^3$ of the $i$-th intruder.
Hence, the resulting strategy for the target-surveillance problem can be chosen by solving the optimization program
\begin{subequations}\label{eq:target_surv}
  \begin{align}
    \min_{\xag[1],\ldots,\xag[\N]} \: & \: \!\! \sum_{i =1}^\N \!\! \left(w_i\norm{\xag[i] - r_i}^2 + \norm{\sum_{i=1}^\N \beta_{i} \xag[i]/\N - r_0}^2\right)
                                        \label{eq:target_cost}\\
    \subj \: & \:
        [r_{i}]_c + \epsilon_c \le [\xag[i]]_c \le [r_0]_c   \quad \text{if } [r_{i}]_c \le [r_0]_c\label{eq:agg_con1}
        \\
        \: & \: [r_0]_c \le [\xag[i]]_c \le [r_{i}]_c  - \epsilon_c \quad \text{if } [r_{i}]_c > [r_0]_c\label{eq:agg_con2}
        \\
        \: & \: c=1, 2, 3, \quad \forall i \in \set.\notag
    \end{align}
\end{subequations}
%
%
%
%
In~\eqref{eq:target_cost}, $w_i, \beta_{i} > 0$ are scaling parameters, while,
in the constraints~\eqref{eq:agg_con1} and~\eqref{eq:agg_con2}, we used
$[\cdot]_c$ to denote the $c$-th component of the vectors and $\epsilon_c > 0$
to denote an additional tolerance.
Problem~\eqref{eq:target_surv} can be cast as an instance of the aggregative optimization problem~\eqref{eq:aggregative_problem} as follows.
First, notice that the cost~\eqref{eq:target_cost} can be split into the sum of $\N$ functions. Each one of them depends
on a local variable $\xag[i]$ and on an aggregated term
\begin{align*}
  \sigma(\xag)&\triangleq \frac{\sum_{i=1}^\N \beta_i \xag[i]}{\N},
\end{align*}
namely, each local aggregation rule reads as $\phi_i(\xag[i]) \triangleq \beta_i \xag[i]$, while each local objective function is defined as
%
\begin{align*}
  \fag[i](\xag[i],\sigma(\xag)) = w_i \norm{\xag[i]-r_i}^2 +\norm{r_0 - \sigma(\xag[])}^2.
\end{align*}
Finally, note that the constraints~\eqref{eq:agg_con1} and~\eqref{eq:agg_con2} are local for each robot. 
Therefore, for each robot $i$, is it possible to define the feasible set $\X[i]$ as
\begin{align*}
  \X[i] \triangleq \Big\{
  \xag[i] \in\real^{\nag[i]} \mid \text{
  \eqref{eq:agg_con1} and~\eqref{eq:agg_con2} 
  are satisfied}
  \Big\}.
\end{align*}

\subsection{\aggropt{2}: Multi-Robot Resource Allocation with Soft Constraints}
\label{sec:aggro2}

Here, we consider a version of the \emph{Multi-Robot Resource Allocation} problem in which we relax the resource allocation constraint and we handle it through a suitable recasting into a distributed aggregative optimization problem.
In this scenario, a team of $\set = \until{\N}$ robots have to cooperatively perform a set of individual tasks by relying on a common, finite resource.
Specifically, we denote by $\xag[i] \in \real$ be the portion of resource allocated to robot $i$ to perform its individual tasks.
Then, we denote with $U_i: \real \to \real$ the utility function used to measure the quality of the service provided by robot $i$ in performing its individual task.
Moreover, it is desirable to not exceed a certain bound $B > 0$ about the aggregate allocated resource $\sum_{i=1}^\N \xag[i]$, so that a penalty function $p: \real \to \real$ depending on $\sum_{i=1}^\N \xag[i] - B$ is added to the cost.
We remark that, here, differently from more standard resource allocation formulations, it is possible to exceed such a bound because it is not modeled through a hard constraint.
Further, we assume that each quantity $\xag[i]$ is constrained to satisfy a bound in the form $\xag[i][{\text{min}}] \leq \xag[i] \leq \xag[i][{\text{max}}]$, for some $\xag[i][{\text{min}}],\xag[i][{\text{max}}] > 0$.
Therefore, we can model this allocation problem through the program
\begin{subequations}\label{eq:resource_agg}
  \begin{align}
    \min_{\xag[1],\ldots,\xag[\N]} \: & \: p\left(\sum_{i=1}^\N \xag[i] - B\right) + \sum_{i =1}^\N U_i(\xag[i])
                                        \label{eq:resource_agg_cost}
    \\
    \subj \: & \: \xag[i][{\text{min}}] \leq \xag[i] \leq \xag[i][{\text{max}}], \forall i \in \set.\label{eq:target_con2}
  \end{align}
\end{subequations}
The objective function in~\eqref{eq:resource_agg_cost} can be split into the sum of $N$ functions depending on the local variable $\xag[i]$ and the aggregated term $\sum_{i=1}^\N \xag[i]$.
Therefore, the aggregative structure~\eqref{eq:aggregative_problem} is recovered by setting each local function as
\begin{align*}
  \fag[i](\xag[i],\sigma(\xag)) = U_i(\xag[i]) + \frac{1}{\N}p(\sigma(\xag) - B),
\end{align*}
where $\sigma(\xag) = \sum_{i=1}^\N \xag[i]$ and, thus, each aggregation rule $\phi_i$ as
\begin{align*}
  \phi_i(\xag[i]) = \N \xag[i].
\end{align*}
Finally, we impose the (local) constraints~\eqref{eq:target_con2} by defining, for each robot $i$, the (local) feasible set $\X[i]$ according to
\begin{align*}
  \X[i] \triangleq \Big\{
  \xag[i] \in\real^{\nag[i]} \mid \text{
  \eqref{eq:target_con2} 
  is satisfied}
  \Big\}.
\end{align*}

%
%
%
%
\section{Distributed Methods for Constrained-Coupled Optimization}
\label{sec:constr_algo}
In this section, we describe two algorithms to address constraint-coupled
optimization problems in the form~\eqref{eq:constr_coupled}, namely the
\emph{Distributed Dual Decomposition}~\cite{falsone2017dual}, and the
\emph{Distributed Primal Decomposition}~\cite{camisa2021distributed}.
The algorithms address convex optimization problems, but suitable extensions can
be designed to handle mixed-integer linear
programs~\cite{vujanic2016decomposition,camisa2021distributed}.
%
%

\subsection{Distributed Dual Decomposition}
Given optimization problems as in~\eqref{eq:constr_coupled}, let us introduce the Lagrangian function $L : \real^{\nag} \times \real^\dcou_+ \to \real$ defined as
\begin{equation*}
 L(\xag,\mu) \triangleq \sum_{i=1}^\N L_i(\xag[i],\mu) \triangleq \sum_{i=1}^\N \left( f_i(\xag[i]) + \mu^\top g_i(\xag[i]) \right),
\end{equation*}
where $\mu\in \real^\dcou_+ $ is the vector of Lagrange multipliers. With this definition at hand,
it is possible to define the (concave) \emph{dual function} as
\begin{align*}
\varphi(\mu) \triangleq \min_{x_1\in X_1, \ldots, x_\N\in X_\N} L(\xag,\mu).
\end{align*}
By exploiting the structure of the Lagrangian, it turns out that $\varphi(\mu) =
\sum_{i=1}^\N \varphi_i(\mu)$ with
\begin{equation}\label{eq:dual_i}
  \varphi_i(\mu) = \min_{x_i\in X_i} L_i(x_i,\mu).
\end{equation}
Thus, we can define the dual problem of~\eqref{eq:constr_coupled} as 
\begin{equation}
  \max_{\mu \geq 0} \sum_{i=1}^\N \varphi_i(\mu). \label{eq:constr_coupled_dual}
\end{equation}
As discussed in~\cite{notarnicola2019constraint}, the dual of a constraint-coupled problem has a consensus optimization structure.
In~\cite{falsone2017dual}, a distributed algorithm based on a distributed proximal on the dual is proposed.
This scheme is reported in Algorithm~\ref{alg:dual_decomp} from the perspective of robot $i$, where $\aij \ge 0$ are the weights of a weighted adjacency matrix associated to the graph and $\stepsize^t > 0$ is the so-called step size at iteration $t$.
%
\begin{algorithm}[!ht]
  \caption{Distributed Dual Decomposition~\cite{falsone2017dual}}
  \begin{algorithmic}[0]
    \State \textbf{Initialization}: $\xag[i][0]\in \X[i]$, $\mu_i^0\in\real^\dcou_+$.
    \For{$t=0, 1, \dots$}
    \begin{align*}
    	v_i^t &= \sum_{j\in\innbrs_i^t}\aij[t] \mu_j^t\\
    	\hxag[i][t+1] &\in \arg\min_{x_i \in \X[i]} L_i(x_i,v_i^t)\\
    	\mu_i^{t+1} &= \arg\max_{\mu_i \geq 0} \big\{ g_i(\hxag[i][t+1])^\top \mu_i- \frac{1}{2\stepsize^t}\Vert\mu_i - v_i^t\Vert^2 \big\}\\
    	\xag[i][t+1] &= \xag[i][t] + \frac{\stepsize^t}{\sum_{r=0}^t \stepsize^r} (\hxag[i][t+1]-\xag[i][t])
    \end{align*}
\EndFor
  \end{algorithmic}
  \label{alg:dual_decomp}
\end{algorithm}


We now briefly explain the main steps in Algorithm~\ref{alg:dual_decomp}.
Each robot $i$ initializes its portion of the solution vector estimate as
$\xag[i][0]\in \X[i]$, and the dual vector estimate
$\mu_i^0\in\real^\dcou_+$ with a feasible solution to
problem~\eqref{eq:constr_coupled_dual}. More details on the initialization
procedure can be found in~\cite{falsone2017dual}.
At every iteration $t\ge 1$, each robot $i$ computes a weighted average
$v_i^t$ of the dual vector based on its own estimate $\mu_i^t$
and on neighbor estimates $\mu_j^t$, $j\in\innbrs_i^t$.
Then, it updates $\hxag[i][t+1]$ by
minimizing the local term $L_i$ of the Lagrangian function evaluated at
$\mu = v_i^t$, and $\mu_i^t$ via a proximal minimization step.
It can be shown that $\hxag[i][t]$ may not converge to an optimal solution $\xag[i][\star]$ to~\eqref{eq:constr_coupled}. Therefore, each robot computes a local candidate solution $\xag[i][t+1]$ as the weighted running average of
$\{\hxag[i][r+1]\}_{r=0}^t$, i.e.,
\begin{equation} \label{eq:xhat_def}
  \xag[i][t+1] = \frac{\sum_{r=0}^t \stepsize^r \hxag[i][r+1]}{\sum_{r=0}^t \stepsize^r}.
\end{equation}
This procedure is often referred to as the primal recovery procedure
of dual decomposition~\cite{nedic2009approximate,chang2014distributed,zhu2011distributed}.
It is worth noting that robots do not exchange any information on their local
estimates $\xag[i][t+1]$. 
Indeed, robots only exchange local dual
estimates. This is an appealing feature in privacy-preserving settings. Indeed, privacy is preserved since there is no trivial disclosure of local decision variables.
Similarly, they do not communicate any information on the local cost or constraints.

\begin{remark}
	In Section~\ref{sec:exp_cco1}, we leverage Algorithm~\ref{alg:dual_decomp} to solve a multi-agent task assignment problem. As pointed out in Section~\ref{sec:task_allocation}, this can be cast as a linear program. 
  Other works in literature, see, e.g.,~\cite{hosseinzadeh2018distributed,richert2015robust,burger2012distributed,burger2011locally}, solve linear problems in a distributed fashion. 
  Algorithm~\ref{alg:dual_decomp} allows robots to find a local optimal sub-vector $\xag[i][\star] \in \real^{\nag[i]}$ (rather than reaching consensus on a common solution $\xag[][\star]$). This is an appealing feature in  those settings (as the task allocation problem) in which the size $\n = \sum_{i=1}^N \nag[i]$ of the decision variable $\xag$ scales up with the number of robots.\oprocend
\end{remark}

We now enforce a set of assumptions needed to provide the main convergence result for Algorithm~\ref{alg:dual_decomp}.
We start from the following standard requirements about the cost functions $\fag[i]$, the constraint functions $g_i$, and the feasible sets $\X[i]$.
\begin{assumption}\label{ass:dualdec_convex}
  For each $i=1,\ldots,\N$, the functions $\fag[i](\cdot)$ and each component of
  $g_i(\cdot)$ are convex. Also, for each $i=1,\ldots,\N$ the sets $\X[i]$ are
  convex and compact.\oprocend
\end{assumption}

\begin{assumption}\label{ass:dualdec_slater}
  There exists $\tilde{x}=\col(\tilde{x}_1,\cdots,\tilde{x}_\N)$, in the
  relative interior of the set $\X$, such that
  $\sum_{i=1}^\N g_i(\tilde{x}_i) \leq 0$ for those components of
  $\sum_{i=1}^\N g_i(\xag[i])$ that are linear in $\xag$, if any, while
  $\sum_{i=1}^\N g_i(\tilde{x}_i) < 0$ for all other components.\oprocend
\end{assumption}

Assumptions~\ref{ass:dualdec_convex} and~\ref{ass:dualdec_slater} are sufficient conditions to have strong duality between problems~\eqref{eq:constr_coupled} and~\eqref{eq:constr_coupled_dual}.

\begin{assumption}\label{ass:dualdec_ck_coefficient}
  $\{\stepsize^t\}_{t \geq 0}$ is a non-increasing sequence of positive reals such that $\stepsize^t \leq \stepsize^r$ for all $t \geq r, r>0$. Also,
  \begin{enumerate}
  \item $\sum_{t=0}^{\infty} \stepsize^t = \infty$,
  \item $\sum_{t=0}^{\infty} (\stepsize^t)^2 < \infty$.\oprocend
  \end{enumerate}
\end{assumption}
Assumption~\ref{ass:dualdec_ck_coefficient} is a standard assumption arising in the case of algorithms employing diminishing step size.
One choice for $\{\stepsize^t\}_{t\ge0}$ matching the conditions required in Assumption~\ref{ass:dualdec_ck_coefficient} is $\stepsize^t = \bar{\stepsize}/(t + 1)$ for some $\bar{\stepsize} > 0$.
\begin{assumption}\label{ass:dualdec_aij_coefficient}
  There exists $\eta \in (0,1)$ such that for all $i,j \in\set$ and all
  $t \geq 0$, $\aij[t] \in [0,1)$, $\aij[t] \geq \eta$, and $\aij[t] > 0$ implies that $\aij[t] \geq \eta$.  
 Moreover, for all $t \geq 0$, $\WeightAdj^t$ is doubly stochastic.\oprocend
\end{assumption}
Assumption~\ref{ass:dualdec_aij_coefficient} is a typical condition required by consensus algorithm and, thus, is inherited by distributed optimization schemes embedding consensus-based mechanisms due to locally reconstruct unavailable global quantites.
\begin{assumption}\label{ass:dualdec_network}
  Let $\EE^t\triangleq\{(j,i)\mid \aij[t]>0\}$ and let
  $\EE^\infty\triangleq\{(j,i)\mid \aij[t]>0\text{ for infinitely many
  }t\}$. Then, the graph $(V,\EE^{\infty})$ is strongly connected, i.e., for any
  two nodes there exists a path of directed edges that connects them. Moreover,
  there exists $T \geq 1$ such that for every $(j,i) \in \EE^{\infty}$, robot
  $i$ receives information from a neighboring robot $j$ at least once every
  consecutive $T$ iterations.\oprocend
\end{assumption} 

Assumption~\ref{ass:dualdec_network} guarantees that any pair of robots communicate directly infinitely often, and the intercommunication interval is bounded. That is, the communication network can be time-varying. Also, the communication graph at a given time instant may not be connected.
The convergence result follows.
\begin{theorem}[\cite{falsone2017dual}] \label{thm:dual_optimality} 
  Consider Distributed Dual Decomposition as given in Algorithm~\ref{alg:dual_decomp}.
  Let Assumptions~\ref{ass:dualdec_convex}-\ref{ass:dualdec_network} hold.
  Let
  $\mu^\star$ denote an optimal solution
  to~\eqref{eq:constr_coupled_dual} and let $\X^\star$ denote the set of minimizers of~\eqref{eq:constr_coupled}. Also, let
  $\xag[][t]=\col{(\xag[1][t],\ldots,\xag[\N][t])}$. Then, 
  for all
  $i=1,\dots,\N$, it stands
  \begin{align*}
    \lim_{t \rightarrow \infty} \|\mu_i^t - \mu^\star\| &= 0\\
    \lim_{t \rightarrow \infty} \texttt{dist}(\xag[][t],\X^\star) &= 0. \eqoprocend
  \end{align*}
\end{theorem}

The distributed dual decomposition technique has been extended in~\cite{tjell2019privacy} to address privacy concerns among agents.
Moreover, the work~\cite{su2022convergence} explores the behavior of distributed dual decomposition in scenarios where inter-agent communication is affected by asynchrony and inexactnesses.
An effective approach to improve the numerical robustness of dual decomposition is the well-known (centralized) algorithm named Alternating Direction Method of Multipliers (ADMM).
A distributed version of ADMM is provided in~\cite{falsone2020tracking}, where a dynamic average consensus protocol is suitably embedded in the parallel version to track missing global information.
Indeed, to solve~\eqref{eq:constr_coupled}, the parallel ADMM would require the knowledge of the coupling constraint violation, which is a global quantity not available at each robot.

\subsection{Distributed Primal Decomposition}
\label{sec:primal}
Primal decomposition is a powerful tool to recast constraint-coupled convex
programs such as~\eqref{eq:constr_coupled} into a master-subproblem
architecture~\cite{silverman1972primal} that is particularly suited for devising distributed strategies.
In the following, we consider coupling
constraints in the form\footnote{We introduced the constraint-coupled setup with
  coupling constraints in the form $\sum_{i=1}^\N g_i(\xag[i]) \le 0$.  The
  additional, constant term $b$ in this section can be easily embedded into the
  functions $g_i$.}  $\sum_{i=1}^\N g_i(\xag[i]) \le b$, with $b\in\real^\dcou$.
The main idea is to interpret the right-hand side vector $b$ as a limited
resource that must be shared among robots. Thus, for all $i \in \set$ we
introduce local \emph{allocation vectors} $y_i \in \real^S$, i.e., $N$ additional decision variables subject to the auxiliary constraint
$\sum_{i=1}^\N y_i = b$. 
Hence, with these new variables at hand, we equivalently rewrite problem~\eqref{eq:constr_coupled} as
\begin{align*}
\begin{split}
      \min_{\substack{\xag[1],\ldots, \xag[\N]\\ y_1,\ldots, y_\N}} \: & \: \sum_{i=1}^\N \fag[i](\xag[i])
    \\
    \subj \: & \: g_i(\xag[i]) \leq y_i, \quad \forall i \in \set\\
    \: & \: \xag[i]\in \X[i], \hspace{.8cm} \forall i \in \set
  \\
  & \: \sum_{i=1}^\N y_i = b.
\end{split}
\end{align*}
In turn, the obtained reformulation can be further rewritten according to a master-subproblem architecture.
That is, robots cooperatively solve a so-called \emph{master problem} in the form
\begin{align}\label{eq:primal_decomp_master}
  \begin{split}
    \min_{y_1,\ldots, y_\N} \: & \: \sum_{i =1}^\N p_i (y_i) 
    \\
    \subj \: & \: \sum_{i=1}^\N y_i = b
    \\
    & \: y_i \in Y_i, \hspace{1cm} \forall i\in\set.
  \end{split}
\end{align}
Here, for each $i\in\set$, the function $\map{p_i}{\real^\dcou}{\real}$ is
defined as the optimal cost of the $i$-th \emph{subproblem}, namely
%
\begin{align}
\begin{split}
  p_i(y_i) \triangleq \: \min_{\xag[i]} \: & \: \fag[i](\xag[i])
  \\
  \subj \: 
  & \: g_i(\xag[i]) \leq y_i,
  \\
  & \: \xag[i] \in \X[i],
\end{split}
\label{eq:primal_decomp_subproblem_conv}
\end{align}
while each $Y_i \subseteq\real^\dcou$ defines the set of variables $y_i$ ensuring feasibility of the corresponding subproblem~\eqref{eq:primal_decomp_subproblem_conv}. 
We note that~\eqref{eq:primal_decomp_subproblem_conv} only uses locally available information.
Thus, given an optimal allocation $(y_1^\star, \ldots, y_\N^\star)$ solving problem~\eqref{eq:primal_decomp_master}, each robot can reconstruct its portion
$\xag[i][\star]$ of an optimal solution to the original problem~\eqref{eq:constr_coupled}
by using its corresponding local allocation $y_i^\star$.

Based on this approach, in the remainder of this section, we focus on a challenging
mixed-integer setup, and we introduce a tailored distributed
primal-decomposition scheme to solve it~\cite{camisa2021distributed}. Then, we
discuss in a remark how the algorithm can be customized to address convex
problems.
%

The distributed optimization algorithm proposed in~\cite{camisa2021distributed}
addresses mixed-integer programs in the form
\begin{align}
  \label{eq:MILP}
  \begin{split}
    \min_{\xag[1],\ldots,\xag[\N]} \: & \: \smallsum_{i =1}^\N c_i^\top \xag[i]
    \\
    \subj \: 
    & \: \smallsum_{i=1}^\N A_i \xag[i] \leq b
    \\
    & \: \xag[i] \in \X[i], i \in \set,
  \end{split}
\end{align}
where $c_i \in \real^{\nag[i]}$ and the decision variables are coupled by $\dcou$ linear constraints,
described by the matrices $A_i \in \real^{\dcou \times \nag[i]}$ %
and the vector $b \in \real^\dcou$.
Here, the decision variable $\xag[i]$ of the generic robot has $\nag[i] = p_i + q_i$
components and the constraint set $\X[i]$ is of the form $\X[i] = P_i \cap (\integer^{p_i} \times \real^{q_i})$,
where $P_i \subset \real^{\nag[i]}$ is a non-empty, compact polyhedron.
Following~\cite{vujanic2016decomposition,falsone2018distributed,falsone2019decentralized,bertsekas2014constrained},
the idea is to solve a convex, relaxed version of~\eqref{eq:MILP}, namely
%
\begin{align}
  \label{eq:LP_restricted}
  \begin{split}
    \min_{\xag[1],\ldots,\xag[N]} \: & \: \smallsum_{i =1}^\N c_i^\top \xag[i]
    \\
    \subj \: 
    & \: \smallsum_{i=1}^\N A_i \xag[i] \leq b - \sigma
    \\
    & \: \xag[i] \in \conv{\X[i]}, i \in \set,
  \end{split}
\end{align}
where $\xag[i] \in \real^{p_i + q_i}$ for all $i \in \set$ and  $\conv{\X[i]}$ denotes the convex hull of $\X[i]$. %
The restriction $\sigma\in\real^\dcou$ is designed to guarantee that~\eqref{eq:LP_restricted} admits a
solution. More details are provided in~\cite{camisa2021distributed}.
%
Once an optimal solution has been found, the optimal solution to~\eqref{eq:MILP} can be reconstructed.
With this reformulation at hand, the idea is to exploit the primal decomposition approach detailed above.
%
%
To this end, robots cooperatively solve a so-called \emph{master problem}, in
the form of~\eqref{eq:primal_decomp_master}, with subproblems 
%
%
%
\begin{align}
  \begin{split}
    p_i(y_i) = \: \min_{\xag[i]} \: & \: c_i^\top \xag[i]
    \\
    \subj \: 
    & \: A_i \xag[i] \leq y_i
    \\
    & \: \xag[i] \in \conv{\X[i]}.
  \end{split}
      \label{eq:primal_decomp_subproblem}
\end{align}
%
The overall method, from the $i$-th robot perspective, is summarized in
Algorithm~\ref{alg:algorithm_primal_milp}, where $\alpha^t$ is the step size at iteraton $t$,
$M > 0$ is an additional tuning parameter, and $\lexmin$ means that the variables $\rho_i, \xi_i$, and $\xag[i]$
are minimized in a lexicographic order~\cite{notarstefano2019distributed}. More in detail, given $v_1,v_2\in\real^n$, $v_1$ is lexicographically larger than $v_2$ ($v_1\lexge v_2$) if $v_1\neq v_2$ and the first non-zero component of $v_1-v_2$ is positive. For a set of vectors $\{v_1,\ldots, v_r\}$, $v_i$ is its lexicographic minimum if $v_j \lexge v_i, \forall j\neq i$. As detailed in~\cite{camisa2021distributed}, this lexicographic minimization step is required to improve the quality of the solution found by each robot.
In Algorithm~\ref{alg:algorithm_primal_milp}, the generic robot $i$ maintains a
local allocation estimate $y_i^t \in \real^S$, and reconstructs a local feasible
solution $\xag[i][T_f]$ to the original problem~\eqref{eq:MILP}.
\begin{algorithm}
  \floatname{algorithm}{Algorithm}
  
  \begin{algorithmic}[0]
    
    \Statex \textbf{Initialization}: $T_f > 0$, $y_{i}^0$ such that $\sum_{i=1}^\N y_i^0 = b - \sigma$ 
   	\For{$t=0, 1, \dots$}
    
    \StatexIndent[0.75]
    \textbf{Compute} $\mu_i^t$ as an optimal Lagrange multiplier of
    \begin{align*} 
      \min_{\xag[i], v_i} \hspace{1.2cm} &\: c_i^\top \xag[i] + M v_i
      \\
      \subj \hspace{0.3cm} 
      \: \mu_i : \: & \: A_i \xag[i] \leq y_i^t + v_i 1
      \\
                                     & \: \xag[i] \in \conv{\X[i]}, \: \: v_i \ge 0,
    \end{align*}
    
    \StatexIndent[0.75]
    \textbf{Receive} $\mu_{j}^t$ from $j\in\nbrs_i$ and update $y_{i}^{t+1}$ with      %
    \begin{align*} 
      y_i^{t+1} = y_i^t + \alpha^t \smallsum_{j \in \nbrs_i} \big( \mu_i^t - \mu_j^t \big),
    \end{align*}
    \EndFor
    \Statex      \textbf{Return} $\xag[i][T_f]$ as optimal solution of
    \begin{align}
      \label{eq:alg_lexmin_MILP}
      \begin{split}      
        \lexmin_{\rho_i, \xi_i, \xag[i]} \: & \: \rho_i
        \\
        \subj \: 
        & \: c_i^\top \xag[i] \leq \xi_i
        \\
        & \: A_i \xag[i] \leq  y_i^{T_f} + \rho_i 1
        \\
        & \: \xag[i] \in \X[i], \: \: \rho_i \ge 0.
      \end{split}
    \end{align}
    \vspace{-0.3cm}    
  \end{algorithmic}
  \caption{Distributed Primal Decomposition for\! MILPs\!\cite{camisa2021distributed}}
  \label{alg:algorithm_primal_milp}
\end{algorithm}

\begin{remark}
  A distributed primal decomposition algorithm for (nonlinear) convex
  constraint-coupled problems in the form of~\eqref{eq:constr_coupled} can be obtained from
  Algorithm~\ref{alg:algorithm_primal_milp} as follows. In the \textsc{Compute}
  step, together with the multiplier $\mu_i^t$, each robot $i$ also stores a
  candidate solution $\xag[i][t]$ to the local subproblem. The sequence
  $\xag[i][t]$ can be shown to converge to a feasible solution
  to~\eqref{eq:constr_coupled}. The final solution step
  of~\eqref{eq:alg_lexmin_MILP} is then not
  required~\cite{camisa2021distributed_b,camisa2019distributed}.\oprocend
\end{remark}

Algorithm~\ref{alg:algorithm_primal_milp} can be shown to achieve finite-time feasibility under the following assumptions.

\begin{assumption}\label{ass:primal_connectivity}
	The communication graph $\GG$ is undirected and connected.
	\oprocend
\end{assumption}
\begin{assumption}
  \label{ass:step size}
  The step size sequence $\{ \alpha^t \}_{t\ge0}$, with each $\alpha^t \ge 0$,
  satisfies $\sum_{t=0}^{\infty} \alpha^t = \infty$,
  $\sum_{t=0}^{\infty} \big( \alpha^t \big)^2 < \infty$.
  \oprocend
\end{assumption}
\begin{assumption}%
  \label{ass:uniqueness}
  For a given $\sigma \ge 0$, the optimal solution of
  problem~\eqref{eq:LP_restricted} is unique.\oprocend
\end{assumption}%
\begin{assumption}%
  \label{ass:slater}
  For a given $\sigma > 0$, there exists a vector $(\hxag[1],\ldots, \hxag[\N])$,
  with $\hxag[i] \in \conv{\X[i]}$ for all $i$, such that
  \begin{align}  
    \zeta & \triangleq \min_{s \in \until{\dcou}} \Big[ b - \sigma -\smallsum_{i=1}^\N A_i \hxag[i] \Big]_s > 0.
            \label{eq:slater_zeta}
  \end{align}
  The cost of $(\hxag[1],\ldots, \hxag[\N])$ is denoted by $J^{\textrm{SL}} = \sum_{i=1}^N c_i^\top \hxag[i]$.
  \oprocend
\end{assumption}
Assumption~\ref{ass:primal_connectivity} requires robots to communicate according to a static network and that the communication should be bi-directional. In Section~\ref{sec:exp_cco3}, we show how this can be achievable also in practical settings using tailored toolboxes.
Assumption~\ref{ass:step size} requires a very typical condition among schemes using diminishing a step size (see also Assumption~\ref{ass:stepsize}), while the technical conditions enforced in Assumptions~\ref{ass:uniqueness} and~\ref{ass:slater} ensure the specific algorithm well-posedeness.
We refer the reader to~\cite{camisa2021distributed} for further details about these requirements.
Let $\sigma^\infty$ denote an \emph{a-priori restriction} computed as described
in~\cite{camisa2021distributed}. Then, the main result can be stated as follows.

\begin{theorem}[\cite{camisa2021distributed}]
  \label{thm:finite_time_feasibility}
Consider Distributed Primal Decomposition for MILPs as given in Algorithm~\ref{alg:algorithm_primal_milp}.
  Let Assumptions~\ref{ass:primal_connectivity},~\ref{ass:step size},~\ref{ass:uniqueness}, and~\ref{ass:slater} hold.
  Let $\sigma^\text{FT} = \sigma^\infty+\delta 1_{S}$ for some
  $\delta > 0$.
  %
  Consider
  the mixed-integer sequence 
  $\{\xag[1][t], \ldots, \xag[\N][t]\}_{t \ge 0}$
  generated by Algorithm~\ref{alg:algorithm_primal_milp} under Assumption~\ref{ass:step size},
  with $\sum_{i=1}^N y_i^0 = b - \sigma^\text{FT}$.
  There exists a sufficiently large time $T_\delta > 0$ such that the vector
  $(\xag[1][t], \ldots, \xag[\N][t])$ is a feasible solution for problem~\eqref{eq:MILP},
  i.e., $\xag[i][t] \in \X[i]$ for all $i \in \until{N}$ and $\sum_{i=1}^N A_i \xag[i][t] \leq b$,
  for all $t \ge T_\delta$.
  \oprocend
\end{theorem}

Primal decomposition has been leveraged to address different distributed  settings. In~\cite{camisa2021distributed_b,camisa2019distributed}, authors propose
distributed primal decomposition schemes to address constraint-coupled convex
problems.
In~\cite{camisa2021distributed}, authors provide numerical simulations showing
that this distributed decomposition scheme allows for the solution of a larger
set of problems and achieves better sub-optimality bounds with respect to
distributed duality decomposition schemes.


\section{Distributed Methods for Aggregative Optimization}
\label{sec:aggregative_algo}
In this section, we describe two schemes to address aggregative optimization
problems in the form~\eqref{eq:aggregative_problem}.
%
The Projected Aggregative Tracking proposed in~\cite{carnevale2021distributed} (as a modified scheme of the one introduced in \cite{li2021distributed_b}) is a
distributed scheme inspired by a projected gradient method.
The Distributed Frank-Wolfe Algorithm with Gradient Tracking proposed
in~\cite{wang2022distributed} is instead inspired by the so-called Franke-Wolfe update.
Recently, in~\cite{carnevale2023nonconvex} (and in the preliminary version~\cite{carnevale2022aggregative}) a distributed feedback law for aggregative problems is designed in the context of feedback optimization.
In~\cite{grontas2022distributed}, the aggregative framework is
addressed via an ADMM-based algorithm.

\subsection{Projected Aggregative Tracking}
\label{sec:pat}
The algorithm in~\cite{carnevale2021distributed} maintains in each
robot $i$, at each iteration $t \in \natural$, an estimate
$\xag[i][t] \in \real^{\nag[i]}$ about the $i$-th block of a solution
$\xag[][\star] \triangleq \col(\xag[1][\star], \dots, \xag[\N][\star]) \in
\real^{\nag}$ to problem~\eqref{eq:aggregative_problem}.
To update such an estimate, one may use the projected gradient method.
In each robot $i$, this method would require the knowledge of the derivative
of $\sum_{j=1}^\N \fag[j](\xag[j],\sigma(\xag))$ with respect to $\xag[i]$
computed in the current configuration
$\xag[][t] \triangleq \col(\xag[1][t],\dots,\xag[\N][t])$.
In light of the chain rule, it reads
\begin{align}
  &\left[\frac{\partial\sum_{j=1}^\N \fag[j](\xag[j],\sigma(\xag))}{\partial \xag[i]}\Bigg|_{\substack{\xag[1] = \xag[1][t]
  \\
  \vdots
  \\
  \xag[\N] = \xag[\N][t]
  }}\right]^\top
  \notag\\
  &= \nabla_1 \fag[i](\xag[i][t],\sigma(\xag[][t])) 
    + \frac{\nabla\phi_i(\xag[i][t])}{\N}\sum_{j=1}^N\nabla_2 f_j(\xag[j][t],\sigma(\xag[][t])).\label{eq:desired_update}
\end{align}
Computing the quantity~\eqref{eq:desired_update} would require the knowledge of
the global quantities $\sigma(\xag[][t])$ and
$\sum_{j=1}^\N\nabla_2 f_j(\xag[j][t],\sigma(\xag[][t]))$.
To overcome this issue, robot $i$ resorts to two auxiliary variables
$s_{i}^t, y_{i}^t \in \real^{\nag[i]}$ called trackers compensating for the
local lack of knowledge.
Specifically, robot $i$ exploits neighboring communication to update the trackers according to two suitable perturbed consensus dynamics.
The whole fully-distributed procedure is summarized in Algorithm~\ref{alg:pat}, where $\aij \ge 0$ are the weights of a weighted adjacency matrix associated to the graph, $\stepsize > 0$ is the step size, while $\delta \in (0,1)$ is a convex combination constant.
\begin{algorithm}[H]
  \begin{algorithmic}
    \State \textbf{Initialization}: $\xag[i][0] \in \X[i]$, $s_{i}^{0} = \phi_{i}(\xag[i][0])$, $y_{i}^{0} = \nabla_2f_{i}(\xag[i][0],s_{i}^0)$
    %
    \For{$t=0, 1, \dots$}
    \begin{align*}
      \tilde{x}_{i}^t &= P_{X_{i}}\big[\xag[i][t] -  \stepsize(\nabla_1 f_{i}(\xag[i][t],s_{i}^{t}) + \nabla\phi_{i}(\xag[i][t])y_{i}^{t})\big]
      \\
      \xag[i][t+1] &= \xag[i][t] + \delta(\tilde{x}_{i}^t-  \xag[i][t] )
      \\
      s_{i}^{t+1} &= \sum_{j=1}^{\N}\aij s_{i}^{t} + \phi_{i}(\xag[i][t+1]) - \phi_{i}(\xag[i][t])
      \\
      y_{i}^{t+1} &= \sum_{j=1}^{\N}\aij y_{j}^{t} + \nabla_2 \fag[i](\xag[i][t+1],s_i^{t+1})
          - \nabla_2 f_{i}(\xag[i][t],s_{i}^{t})
    \end{align*}
    \EndFor
  \end{algorithmic}
  \caption{Projected Aggregative Tracking (Robot $i$)~\cite{carnevale2021distributed}}
  \label{alg:pat}
\end{algorithm}
The main convergence result for Algorithm~\ref{alg:pat} requires the following assumptions where, for the sake of readability, we adopt
$G_1 :\real^{\nag} \times \real^{\N d} \to \real^{\nag}$, 
$G_2 :\real^{\nag} \times \real^{\N d} \to \real^{\N d}$, 
and $G :\real^{\nag} \times \real^{\N d} \to \real^{\nag}$ defined as 
\begin{align*}
  G_1(x,s) &\triangleq \begin{bmatrix}
    \nabla_1 \fag[1](\xag[1],s_1)
    \\
    \vdots
    \\
    \nabla_1 \fag[\N](\xag[\N],s_\N)
  \end{bmatrix}\!\!, G_2(x,s) \triangleq \begin{bmatrix}
    \nabla_2 \fag[1](\xag[1],s_1)
    \\
    \vdots
    \\
    \nabla_2 \fag[1](\xag[\N],s_\N)
  \end{bmatrix}
  \\
  G(\xag,s) &\triangleq G_1(\xag,s) + \nabla\phi(\xag)G_2(\xag,s),
\end{align*}
where $x \triangleq \col(\xag[1],\dots,\xag[\N])$ and $s \triangleq \col(s_1,\dots,s_\N)$ with $\xag[i] \in \real^{\nag[i]}$ and $s_i \in \real^d$ for all $i \in \set$.

The following assumption states the network connectivity properties required in~\cite{carnevale2021distributed} for the execution of Algorithm~\ref{alg:pat}.
\begin{assumption}\label{ass:network}
  The graph $\mathcal{G}$ is undirected and connected and $\WeightAdj$ is doubly stochastic.
  \oprocend
\end{assumption}
\begin{assumption}\label{ass:convexity}
 For all $i \in \set$, $\X[i] \subseteq \real^{\nag[i]}$ is nonempty, closed, and convex, while the global objective function $f(\xag,\sigma(\xag))$ is $\mu$-strongly convex.\footnote{A function $\fag:\real^n\to\real$ is said to be $\mu$-strongly convex, for $\mu>0$, if $\forall x_1,x_2\in\real^n$, $\lambda\in[0,1]$
  it holds $f(\lambda x_2+(1-\lambda)x_1)\leq \lambda f(x_2)+(1-\lambda)f(x_1)-\frac{\mu}{2}\lambda(1-\lambda)\Vert x_1-x_2\Vert^2$.}. \oprocend
\end{assumption}
Assumption~\ref{ass:convexity} ensures that problem~\eqref{eq:aggregative_problem} has a unique solution $\xag[][\star] \in \real^n$.
The following assumptions, instead, generalize in the aggregative setup the typical Lipschitz properties needed when employing gradient-based optimization schemes.
\begin{assumption}\label{ass:lipschitz}
  The function $f(\xag,\sigma(\xag))$ is differentiable with $L_1$-Lipschitz continuous gradients,
  and 
  $G$, $G_2$ are Lipschitz continuous 
  with constants $L_1, L_2 > 0$, respectively.
  For all $i \in \set$, the aggregation function $\phi_i$ is differentiable
  and $L_3$-Lipschitz continuous. 
  \oprocend
\end{assumption}
The following theorem guarantees the linear convergence of Algorithm~\ref{alg:pat} to the unique solution $\xag[][\star]$ to problem~\eqref{eq:aggregative_problem}. 
\begin{theorem}[\cite{carnevale2021distributed}]\label{th:pat}
  Consider Projected Aggregative Tracking as given in
  Algorithm~\ref{alg:pat}.
  Let Assumptions~\ref{ass:network},~\ref{ass:convexity}, and~\ref{ass:lipschitz} hold. 
  Then there exist $\lambda, \bar{\delta} > 0$ and $\tilde{\rho} \in (0,1)$ such that, if $\stepsize \leq \frac{1}{L_1}$ and $\delta \in (0,\bar{\delta})$, it holds
  \begin{align*}
    &f(\xag[][t],\sigma(\xag[][t])) - \fag(\xag[][\star],\sigma(\xag[][\star])) 
    \\
    &\le \tilde{\rho}^{2t}\frac{L_1 \lambda^2}{2} \norm{\begin{bmatrix}\norm{\xag[][0] - \xag[][\star]}
        \\
        \norm{s^0 - \sigma(\xag[][0])}
        \\
        \norm{y^0 - \sum_{j=1}^\N \nabla_2 \fag[j](\xag[j][0],\sigma(\xag[][0]))/N}
      \end{bmatrix}}^2,
  \end{align*}
  where $s^0 \triangleq \col(s_1^0,\dots,s_\N^0)$ and $y^0 \triangleq \col(y_1^0,\dots,y_\N^0)$.\oprocend
\end{theorem}
Theorem~\ref{th:pat} ensures that Algorithm~\ref{alg:pat} linearly converges toward the problem solution $\xag[][\star]$.
It is worth noting that, in the case of more weaker problem assumptions, the algorithms' convergence guarantees will be weaker too.
For instance, the work~\cite{carnevale2023nonconvex} deals with nonconvex aggregative problems and accordingly guarantees only asymptotic convergence toward generic stationary points of the problem.

A first, pioneering version of Algorithm~\ref{alg:pat} was proposed
in~\cite{li2021distributed_a} for the static, unconstrained setting. A first
version for the online, constrained setting was introduced
in~\cite{li2021distributed_b}, whose continuous-time counterpart has been
extended in~\cite{chen2023distributed} to deal with quantized communication.
In~\cite{carnevale2022learning}, Algorithm~\ref{alg:pat} has been combined with
a Recursive Least Squares mechanism to learn unknown parts of the objective
functions in a \emph{personalized} setting.
It is worth noting that Algorithm~\ref{alg:pat} achieves linear convergence
(cf. Theorem~\ref{th:pat}) in solving problem~\eqref{eq:aggregative_problem}.
As for the unconstrained setting, the linear convergence is also 
achieved by the
distributed method proposed in~\cite{li2021distributed_a}, 
its extension
in~\cite{wang2022distributed} dealing with quantized communication, 
and its accelerated versions proposed in~\cite{liu2023accelerated}.

\subsection{Distributed Frank-Wolfe Algorithm with Gradient Tracking}
\label{sec:FW}

The algorithm in~\cite{wang2022distributed} is inspired by the so-called Franke-Wolfe method.
The main difference between the projected gradient method and the Franke-Wolfe
method lies in the fact that the latter, at the cost of losing linear
convergence, requires a lower computational cost.
When applied to problem~\eqref{eq:aggregative_problem}, in each robot $i$, the Franke-Wolfe method reads as
\begin{subequations}\label{eq:desired_FW}
  \begin{align}
    z^t_i &= \argmin_{z_i \in \X}(d_i^t)^\top z_i
    \\
    \xag[i][t+1] &= (1 - \stepsize^t)\xag[i][t] + \stepsize^t z_i^t,
  \end{align}
\end{subequations}
where $\stepsize^t$ is the time-varying step size at iteration $t \in \natural$, while, in order to ease the notation, we introduced $d_i^t$ to denote the derivative of $f(\cdot,\sigma(\cdot))$ computed at $\xag[][t]$, namely 
\begin{align*}
  d_i^t \triangleq \nabla_1 \fag[i](\xag[i][t],\sigma(\xag[][t])) 
  + \frac{\nabla\phi_i(\xag[i][t])}{\N}\sum_{j=1}^\N\nabla_2 f_j(\xag[j][t],\sigma(\xag[][t])).
\end{align*}
As in Section~\ref{sec:pat}, the presence of global, unavailable quantities required to compute $d_i^t$ makes the update~\eqref{eq:desired_FW} not implementable in a distributed fashion.
For this reason, the update~\eqref{eq:desired_FW} is modified and suitably interlaced with two trackers $s_i^t$ and $y_i^t$ having the same role and dynamics as in Section~\ref{sec:pat}.
The whole procedure is named Distributed Frank-Wolfe Algorithm with Gradient Tracking and is summarized in Algorithm~\ref{alg:FW}.
In~\cite{wang2022distributed}, the algorithm has been studied to deal with time-varying graphs $\mathcal{G}^t$.
Indeed, we notice that the trackers' updates depend on time-varying weights $\aij[t]$ corresponding to a time-varying adjacency matrix $\WeightAdj^t$ associated to $\mathcal{G}^t$.
\begin{algorithm}[H]
  \begin{algorithmic}
    \State \textbf{Initialization}: $\xag[i][0] \in \X[i]$, $ s_{i}^{0} = \phi_{i}(\xag[i][0])$, $y_{i}^{0} = \nabla_2f_{i}(\xag[i][0],s_{i}^0)$
    %
    \For{$t = 0, 1, \dots$}
    \begin{align*}
      z^t_i &= \argmin_{z_i \in \X[i]}(\nabla_1 \fag[i](\xag[i][t],s_{i}^{t}) + \nabla\phi_{i}(\xag[i][t])y_{i}^{t})^\top z_i
      \\
      \xag[i][t+1] &= (1 - \stepsize^t)\xag[i][t] + \stepsize^t z_i^t
      \\
      s_{i}^{t+1} &= \sum_{j=1}^{\N}\aij[t] s_{i}^{t} + \phi_{i}(\xag[i][t+1]) - \phi_{i}(\xag[i][t])
      \\
      y_{i}^{t+1} &= \sum_{j=1}^{\N}\aij[t] y_{j}^{t} + \nabla_2 f_{i}(\xag[i][t+1],s_i^{t+1})
          - \nabla_2 \fag[i](\xag[i][t],s_{i}^{t})
    \end{align*}
    \EndFor
  \end{algorithmic}
  \caption{Distributed Frank-Wolfe Algorithm with Gradient Tracking (Robot $i$)~\cite{wang2022distributed}}
  \label{alg:FW}
\end{algorithm}
We state suitable assumptions to provide the convergence result related to Algorithm~\ref{alg:FW}.
\begin{assumption}\label{ass:objective_FW}
  For each robot $i \in \set$, the feasible set $\X[i]$ is closed, convex, and compact, while the global objective function $\fag$ is convex and differentiable over $\X$.\oprocend
\end{assumption}
\begin{assumption}\label{ass:network_FW}
  $\WeightAdj^t$ is doubly stochastic for all $t \in \natural$. 
  Moreover, there exists a constant $0 < \eta < 1$ such that $a_{ij}^t \ge \eta$ for all $j \in \mathcal{N}_i$, $i\in\set$, and $t \in \natural$.
  %
  Further, there exists $B \in \natural$ such that the union graph $\cup_{\tau=t}^{t + B}\mathcal{G}^\tau$ is strongly connected for all $t \in \natural$.
  \oprocend
\end{assumption}
\begin{assumption}\label{ass:stepsize}
  For all $t \in \natural$, it holds $0 \leq \stepsize^{t + 1} \leq \stepsize^t \leq 1$, $\sum_{k=0}^\infty \stepsize^t = \infty$, and
  $\sum_{t = 0}^\infty (\stepsize^t)^2 < \infty$.\oprocend
\end{assumption}
Differently to Assumption~\ref{ass:convexity}, here the strong convexity of $\sum_{i=1}^N \fag[i](\xag[i],\sigma(\xag))$ is not required (cf. Assumption~\ref{ass:objective_FW}).
Moreover, also the network connectivity properties stated in Assumption~\ref{ass:network_FW} are weaker than those in Assumption~\ref{ass:network}.
As a drawback, Algorithm~\ref{alg:FW} needs to employ a diminishing step size $\stepsize^t$ following the rule formalized in Assumption~\ref{ass:stepsize}.
With these assumptions at hand, we are ready to provide the convergence properties of Algorithm~\ref{alg:FW} toward the optimal cost $f^\star$ to problem~\eqref{eq:aggregative_problem} (whose existence is guaranteed by Assumption~\ref{ass:objective_FW}).
\begin{theorem}[\cite{wang2022distributed}]\label{th:FW}
  Consider Distributed Frank-Wolfe Algorithm with Gradient Tracking as given in Algorithm~\ref{alg:FW}.
  Let Assumptions~\ref{ass:lipschitz},~\ref{ass:objective_FW},~\ref{ass:network_FW}, and~\ref{ass:stepsize} hold.
  Then, it holds 
  \begin{align*}
    \lim_{t \to \infty}f(\xag[][t],\sigma(\xag[][t])) = \fag[][\star].\eqoprocend
  \end{align*}
\end{theorem}

As one may expect, Theorem~\ref{th:FW} does not ensure a linear rate for the convergence of Algorithm~\ref{alg:FW}.
This fact is not surprising since we remark that, in general, Franke-Wolfe updates do not exhibit such a property even with a constant step size.

\subsection{Dual Consensus ADMM}
The work in~\cite{grontas2022distributed} proposed an ADMM-based method, inspired by the work in~\cite{mateos2010distributed}, to solve a simplified class of aggregative optimization problems~\eqref{eq:aggregative_problem} that read as
\begin{align}
  \begin{split}
    \min_{\xag[1],\ldots \xag[\N]} \: & \: \sum_{i =1}^\N \fag[i](\xag[i])+g(\sigma(x)),
    \\
    \subj \: & \: \xag[i] \in \X[i], \forall i \in \set
    \\
    \:  & \: \sigma(x) \triangleq \frac{\sum_{i =1}^\N Q_i\xag[i]}{N},
  \end{split}
	\label{eq:aggregative_grontas}
\end{align}
with $\X[i]=\{\xag[i]\in\real^{\nag[i]}\mid E_i\xag[i]=e_i, M_i\xag[i]\leq m_i \}$ with $Q\in\real^{m\times\nag[i]}$ and $E_i,e_i,M_i,m_i$ matrices and vector of appropriate size. The procedure proposed in~\cite{grontas2022distributed} allows for the solution of~\eqref{eq:aggregative_grontas} by considering a dual version of~\eqref{eq:aggregative_grontas} and then recovering the primal solution.
In particular, the dual problem of~\eqref{eq:aggregative_grontas} can be shown to be
\begin{align}
    \max_{y} \: & \: -\sum_{i =1}^\N \fag[i][\star](-Q_i^\top y)+ \frac{1}{\N}g^\star(y),
	\label{eq:dual_aggregative_grontas}
\end{align}
where $\fag[i][\star]\triangleq\inf_x\{\fag[i](\xag[i])+y^\top Q_i\xag[i]\}$ and $g^\star\triangleq\inf_w\{g(w)-y^\top w\}$ are the conjugate functions of $\fag[i]$ and $g$, respectively.
The scheme is outlined in Algorithm~\ref{alg:ADMM_aggr}, where $\Prox_g^{1/(N\xi)}$, is the proximal operator and $d_i=|\innbrs_i|$. 
We refer the reader to~\cite{grontas2022distributed,mateos2010distributed} for a rigorous definition.
 \begin{algorithm}[H]
  \begin{algorithmic}
    \State \textbf{Initialization}: $\rho,\xi>0$, $p_i^0=0_m$, $y_i^0, z_i^0,s_i^0\in\real^m$
    \For{$t = 0, 1, \dots$}
    \begin{align}
      p_i^{t+1} &= p_i^t +\rho\sum_{j\in\innbrs_i}(y_i^t-y_j^t)
      \\
      s_i^{t+1} &= s_i^t +\xi(y_i^t-z_i^t)
      \\
      r_i^{t+1} &= \rho\sum_{j\in\innbrs_i}(y_i^t+y_j^t) + \xi z_i^t - p_i^{t+1} - s_i^{t+1}\label{eq:ritk}
    \end{align}
    \State Compute the optimal value $\xag[i][t+1]$ to
      \begin{align}
     \label{eq:admm_step_8}
      \begin{split}
	\argmin_{\xag[i]}& \left\{\fag[i](\xag[i]) + \frac{1}{2(\xi+2\rho d_i)} \Vert Q_i\xag[i] + r_i^{t+1} \Vert^2\right\}
	\end{split}
    \end{align}
    \State Update $\xag[i][t+1]$ solving
    \begin{align}
      \label{eq:admm_step_9}
      \begin{split}
	\argmin_{\xag[i]}&\left\{\fag[i](\xag[i]) + \frac{1}{2(\xi+2\rho d_i)} \Vert Q_i\xag[i] + r_i^{t+1} \Vert^2\right\}
  \\
	\subj &\quad \Vert \xag[i] \Vert_{\infty} \leq M_i
	\end{split}
    \end{align}
    If the solution is unfeasible/suboptimal, update $M_i = 2 M_i$
    \State Once an optimal solution has been found, update
    \begin{align*}
      y_i^{t+1} &= \frac{1}{2(\xi+2\rho d_i)} ( Q_i\xag[i] + r_i^{t+1} )\\
      z_i^{t+1} &= \frac{s_i^{t+1}}{\xi} + y_i^{t+1} - \frac{1}{\N\xi}\Prox_g^{1/(\N\xi)}(N s_i^{t+1}+\xi y_i^{t+1})
%
      \end{align*}
    \EndFor
  \end{algorithmic}
  \caption{Dual Consensus ADMM (Robot $i$)~\cite{grontas2022distributed}}
  \label{alg:ADMM_aggr}
\end{algorithm}

%
The well-posedness of the minimizations required in the execution of each iteration of Algorithm~\ref{alg:ADMM_aggr} require stronger assumptions than those enforced to provide the convergence properties of Algorithm~\ref{alg:pat} and~\ref{alg:FW}.
We state these assumptions as follows and, then, we provide the convergence result related to Algorithm~\ref{alg:ADMM_aggr}.

\begin{assumption}\label{ass:proper_func}
The following holds

$1)$ The functions $\fag[i], \forall i\in\set$, and $g$ are proper, closed, and convex.

$2)$ A primal-dual solution exists and strong duality holds.

$3)$ The graph $\GG$ is connected and undirected.
\oprocend
\end{assumption}
\begin{assumption}\label{ass:ritk}
Consider $r_i^{t+1}$ as defined in~\eqref{eq:ritk}.
For all $i\in\set$ and for all $\tilde{r}\in \real^m$, there exist $R, M_i>0$ finite such that $\Vert r_i^{t+1}-\tilde r \Vert \leq R $ implies that the solution set of~\eqref{eq:admm_step_8} intersects the set $\{\xag[i]\in\real^{\nag[i]} \mid \Vert \xag[i] \Vert_\infty \leq M_i\}$.
\oprocend
\end{assumption}

With these assumptions at hand, the main result follows.

\begin{theorem}[\cite{grontas2022distributed}]\label{th:grontas}
Consider Dual Consensus ADMM as given in Algorithm~\ref{alg:ADMM_aggr}.
Let Assumptions~\ref{ass:proper_func} and~\ref{ass:ritk} hold.
The iterates $(\{y_i^t\}_{i\in\set})_{t\in\natural}$, $(\{z_i^t\}_{i\in\set})_{t\in\natural}$ of Algorithm~\ref{alg:ADMM_aggr} converge to a maximizer of the dual~\eqref{eq:dual_aggregative_grontas}, while $(\{\xag[i][t]\}_{i\in\set})_{t\in\natural}$ converges to the set of minimizers of the primal~\eqref{eq:aggregative_grontas}.\oprocend
\end{theorem}

Although Algorithm~\ref{alg:ADMM_aggr} requires a simplified problem structure (cf.~\eqref{eq:aggregative_grontas}) and a higher computational burden due to the minimizations required in each algorithm iteration, we highlight that Theorem~\ref{th:grontas} does not guarantee a linear convergence rate for Algorithm~\ref{alg:ADMM_aggr}.
However, since the versatility and robustness properties of ADMM-based strategies have been already studied for other setups, see, e.g.,~\cite{bastianello2018distributed,bastianello2020asynchronous} for the consensus optimization setup, and~\cite{falsone2020tracking} for the constraint-coupled one, Algorithm~\ref{alg:ADMM_aggr} may be a good starting point for future developments about aggregative scenarios considering non-ideality issues often encountered in realistic multi-robot setups (e.g., communication over imperfect networks and asynchrony among the robots).

\section{Toolboxes and Experiments}
\label{sec:toolbox}
\begin{figure}[ht]
  \centering
  \includegraphics[width=\columnwidth]{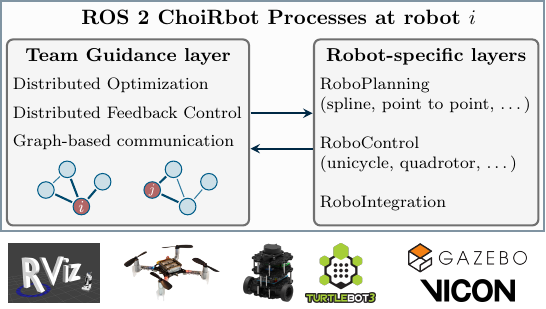}
  \caption{\textsc{ChoiRbot} architecture. Each robot is implemented as a set of independent ROS~2 nodes that implement
      the distributed optimization algorithms as well as low-level planning and control routines.}
  \label{fig:choirbot_architecture}
\end{figure}

To further highlight the potential of the considered frameworks in multi-robot
applications, in this section we provide simulations and experiments on teams of robots for
the proposed theoretical settings. All the code is available on GitHub\footnote{https://github.com/OPT4SMART/multirobot-tutorial}. 
A website with tailored tutorials is also available on GitHub Pages\footnote{https://opt4smart.github.io/multirobot-tutorial/}.
The implementation of the distributed algorithms has been performed using the 
DISROPT Python package~\cite{farina2020disropt}. This framework allows users to implement
distributed algorithms from the perspective of the generic robot in the network and provides
API to exchange data according to user-defined graphs.
The implementation on the robotic platform instead is performed using \textsc{ChoiRbot}~\cite{testa2021choirbot},
and \textsc{CrazyChoir}~\cite{pichierri2023crazychoir}, two novel ROS~2 frameworks tailored for multi-robot applications.
Specifically, \textsc{ChoiRbot} is a general-purpose framework allowing for the implementation of distributed optimization
and control schemes on generic robots, whereas \textsc{CrazyChoir} is tailored to manage Crazyflie nano-quadrotor fleets.
Both frameworks are based on the ROS~2 framework.
Each robot is modeled as a cyber-physical agent and consists
of three interacting layers:
distributed optimization layer, trajectory planning layer, and low-level control layer.
For each robot, each layer is handled by a different ROS~2 process. As for the distributed optimization
layer, the two frameworks support the implementation of algorithms using DISROPT, and implement 
inter-robot communication using the inter-process communication stack of ROS~2.
To better clarify this structure, we report the \textsc{ChoiRbot} architecture in Figure~\ref{fig:choirbot_architecture}. 
  Here, it is possible to see how these toolboxes allow for the implementation of distributed algorithms in a flexible and
  scalable fashion. 
  It is worth noting that, in these frameworks, low-level planning and control are independent of the implementation of distributed algorithms.
  In this way, users can also resort to existing low-level control routines to actuate the robots.
  Moreover, the toolboxes allow for seamless integration with external simulators, e.g., Gazebo~\cite{koenig2004design} or Webots~\cite{michel2004cyberbotics}, and allow users to switch from simulation to experiments by changing a few lines of code.
Indeed, these simulators allow for a realistic representation of the robots leveraging CAD tools, thus resulting in a realistic
  simulation which includes motor dynamics, frictions, and onboard sensors.
In the ROS~2 framework, communication is implemented via a publish-subscribe protocol, enabling message exchange among different computing units connected across the network. Leveraging this feature, the experiments are performed over a real WiFi network, thus taking into account delays and asynchronous (or lossy) communications over a given, user-defined communication graph. More in detail, in the considered experiments we assumed robots communicate according to a static, undirected graph. Graphs are generated as Erd\H{o}s-R\'{e}nyi random graphs with edge probability $0.2$.
 We leveraged \textsc{ChoiRbot} and \textsc{CrazyChoir} functionalities to implement this kind of communication over WiFi networks.

We conclude this introductory part by highlighting that works in the literature often neglect implementations details on how communication and cooperation among robots is implemented or resort to all-to-all broadcasts architecture only simulating the desired distributed paradigm.

\subsection{Experiments for \constrcoup{1}}
\label{sec:exp_cco1}

In the following, we provide Gazebo simulations for the multi-agent multi-task assignment setting as in Section~\ref{sec:task_allocation}. We first consider a generic task assignment scenario in which a team of robots has to serve a set of tasks while minimizing the total travelled distance. Then, we consider the coverage scenario proposed in~\cite{chung2022distributed}.

\begin{figure}[ht]
  \centering
  \includegraphics[width=.98\columnwidth]{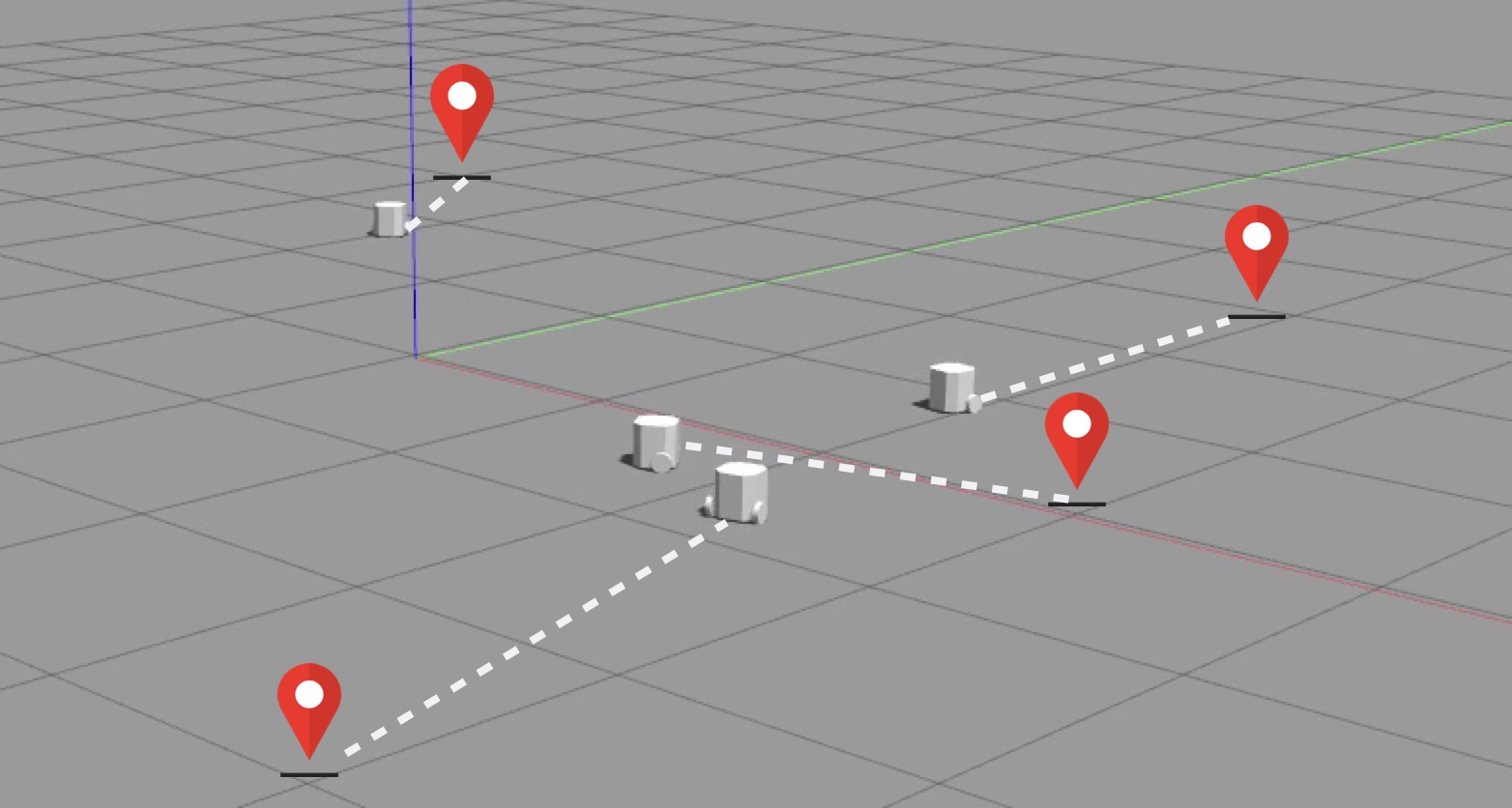}
  \caption{Snapshot from the dynamic task assignment simulation.
    Robots move in order to reach their designed tasks (red markers).}
  \label{fig:exp}
\end{figure}

For our first simulation, we consider a team of $4$ TurtleBot 3 Burger mobile robots tackling a task allocation problem as in Section~\ref{sec:task_allocation} by using the Dual Decomposition scheme in Algorithm~\ref{alg:dual_decomp}. 
In the considered scenario, robots have to fulfill $4$ tasks, randomly scattered in the environment. We assume that robots have access to a shared memory that contains all the needed data. We highlight that this memory does not implement any computing procedure but solely stores the details of the tasks.
Robots start their mission at positions $p_i$. For the sake of simplicity, we assume that a task is accomplished once the allocated robot reaches its position $p_j^T$. The cost value $c_{ij}$ in~\eqref{eq:task_assignment} is $c_{ij} = \Vert p_i - p_j^T\Vert^2$.
%
%
Following the setup described in~\cite{testa2021choirbot}, and to showcase how distributed optimization can be employed also in dynamic settings, we consider the
scenario in which more tasks arrive during time.  We assume that a new task is
revealed as soon as robots complete an already-known task. As soon as robots
receive information on new tasks, they restart the optimization procedure and
evaluate a novel, optimal allocation. This re-optimization approach has been also considered in~\cite{testa2021generalized,testa2021choirbot,karaman2008large}. Although tailored optimization approaches have been proposed to address dynamic problems, this is beyond the scope of this tutorial.
As soon as robots finish the optimization,
they start moving toward their designated task. To this end, we have a
proportional control scheme.  To avoid collisions, the control inputs are fed
into a collision avoidance scheme using a barrier function as
in~\cite{wilson2020robotarium}.
In Figure~\ref{fig:exp}, we provide a snapshot of the simulation\footnote{A video of the simulation is available at~\url{https://youtu.be/9YzbdmIgCYg}.}.

\begin{figure}[!htpb]
\centering
  \includegraphics[width=.495\columnwidth]{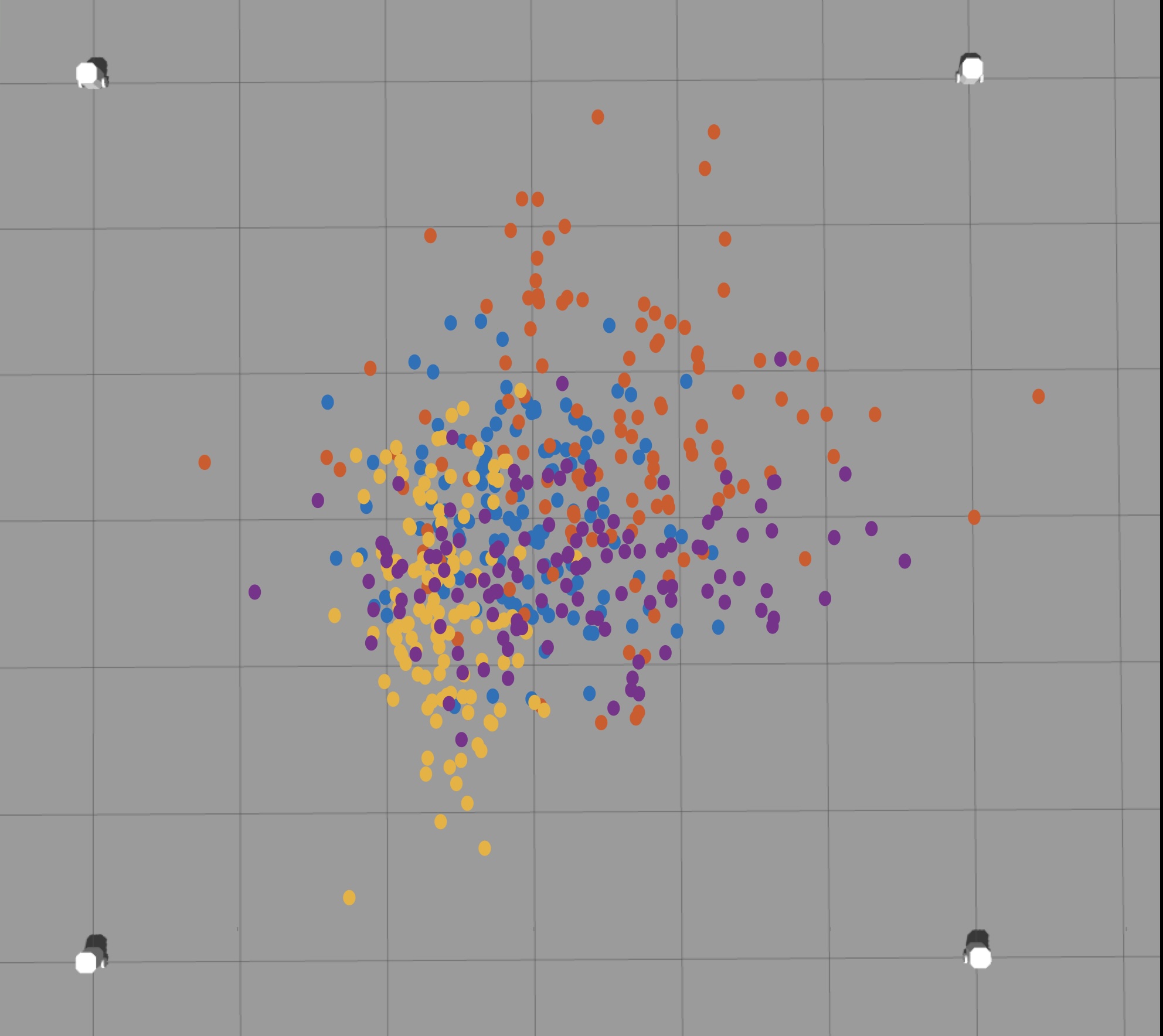}
  \caption{Distribution of the targets. Each color represents the sample available to each robot. White robots are posed in the environment waiting to be deployed.}
  \label{fig:target_distr}
\end{figure}

Now, we implement the setting in~\cite{chung2022distributed} as detailed in Section~\ref{sec:task_allocation}. We consider a network of $4$ mobile robots that have to be deployed to
provide a service to a group of targets densely scattered in the environment. Each robot is aware of only part of the targets. That is, we generated the targets according to $4$ different Gaussian distributions with the following parameters: $\mu_1=\begin{bmatrix}
0 & 0	
\end{bmatrix}, \mu_2=\begin{bmatrix}
2/3 & 2/3	
\end{bmatrix}, \mu_3=\begin{bmatrix}
-2/3 & 0-2/3
\end{bmatrix}, \mu_4=\begin{bmatrix}
1/3 & -1/3	
\end{bmatrix}$ and $\sigma_1=\begin{bmatrix}
	0.3 & 0\\0 & 0.4
\end{bmatrix}, \sigma_2=\begin{bmatrix}
	0.6 & 0\\0 & 0.8
\end{bmatrix}, \sigma_3=\begin{bmatrix}
	0.1 & 0\\0 & 0.5
\end{bmatrix}, \sigma_4=\begin{bmatrix}
	0.8 & 0\\0 & 0.2
\end{bmatrix}$.
 In Figure~\ref{fig:target_distr}, we show a snapshot from a Gazebo simulation with the target distribution and the information available to each robot.

\begin{figure}[!htpb]
\centering
  \includegraphics[width=.495\columnwidth]{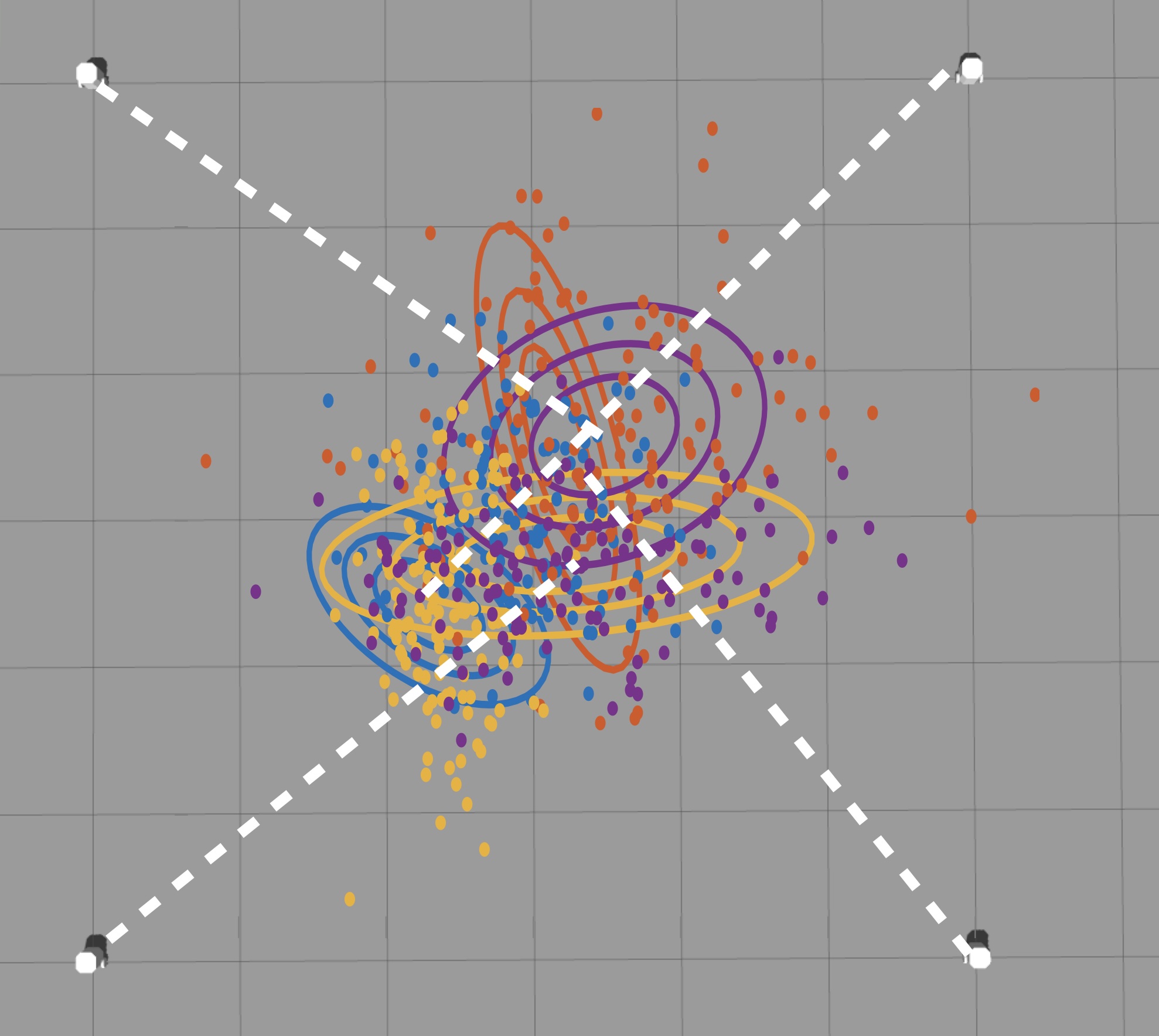}
  \caption{Clusters estimated by the robots. Each color indicates the data available to each robot and the allocated cluster. Dashed lines represent the trajectory to be tracked by each robot to reach the cluster.}
  \label{fig:target_estim}
\end{figure}

The Gaussian Mixture of each robot is initialized with a set of random means and random diagonal covariance matrices. These data have been generated according to a uniform distribution. We implemented the algorithm in~\cite{chung2022distributed} in order to make robots estimate the target distributions.
We assumed robots have the following QoS profiles. We set  $\omega_i=\frac{1}{4}$ for each $i\in\set$. As for the covariance matrix, they are in the form $\Sigma_i=R(\theta_i)\Lambda_i R(\theta_i)^\top$ with $\Lambda_1=\begin{bmatrix}
	1 & 0\\0 & 2
\end{bmatrix}, \Lambda_2=\begin{bmatrix}
	3 & 0\\0 & 1.5
\end{bmatrix}, \Lambda_3=\begin{bmatrix}
 1.5 & 0\\0 & 2.5
\end{bmatrix}, \Lambda_4=\begin{bmatrix}
	1 & 0\\0 & 2
\end{bmatrix}$.
 We implemented the procedure in~\cite{chung2022distributed} to evaluate the cost coefficients for problem~\eqref{eq:task_assignment} and we solved the distributed task allocation problem as in the previous setting. Let   $\xag[i][\star]$ denote the optimal solution of the task assignment problem for robot $i$. The entry $\xag[ij][\star]=1$ denotes that robot $i$ has been assigned to cluster $j$. In Figure~\ref{fig:target_estim}, we show the $j$-th Gaussian estimated by the $i$-th robot for $j$ such that $\xag[ij][\star]=1$. 

With this solution at hand, each robot will be deployed at a position $\mu_{i,j}$. Since the deployment follows a similar evolution to the one shown in the previous example, we omit it.

\subsection{Simulations for \constrcoup{2}}
\label{sec:exp_cco2}

\begin{figure}[!htpb]
  \includegraphics[width=.495\columnwidth]{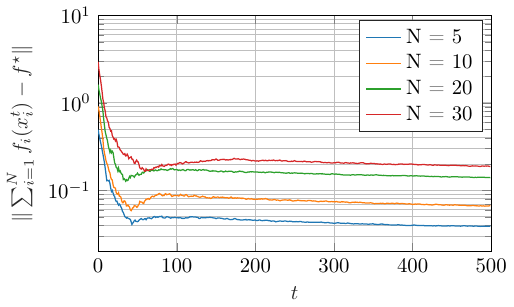}
  \includegraphics[width=.495\columnwidth]{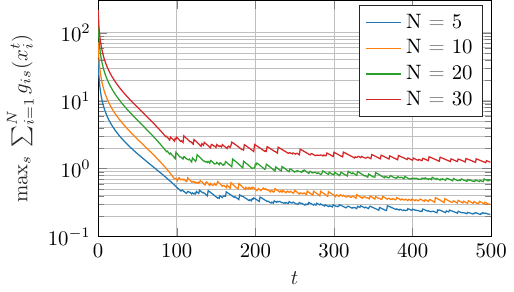}
  \vspace{-.7cm}
  \caption{Cost error and constraint violation changing the number of robots.}
  \label{fig:sim_pev}
\end{figure}

In this section, we provide numerical simulations for the constraint-coupled optimization setup described in Section~\ref{sec:pev}.
The distributed algorithm has been implemented using DISROPT from the perspective of the $i$-th robot as in Algorithm~\ref{alg:dual_decomp}.
In DISROPT, each robot is simulated using an isolated process. Different processes communicate according to a user-defined graph using the MPI (Message Passing Interface) protocol. We run a Monte Carlo simulation for a different number of robots. That is, we considered the cases with $\N=5,10,20,30$. 
For each case, we run $20$ different simulations with varying parameters. More in detail, 
we assumed a charging period of $T=24$ time slots, with each time slot having a duration of $20$ minutes. All the other quantities in~\eqref{eq:EV_primal_problem} have been generated according to~\cite{vujanic2016decomposition}. During each simulation, we run the algorithm for $500$ iterations. Then, we evaluated the maximum for each iteration over the $20$ simulations.
Results are provided in Figure~\ref{fig:sim_pev}. Here, we provide the cost error and the violation of the coupling constraints. As expected from theory, robot estimates converge towards the optimal value, and the violation of the coupling constraints approaches $0$.

\subsection{Experiments for \constrcoup{3}}\label{sec:exp_cco3}
\begin{figure}[htbp]
  \centering
  %
  \includegraphics[width=.9\columnwidth]{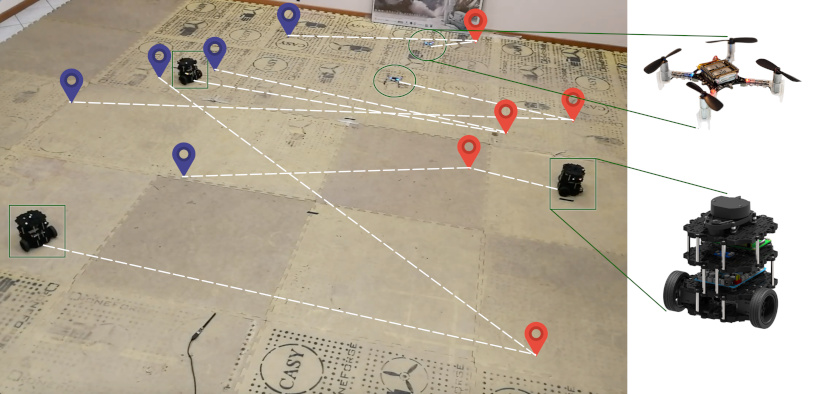}
  \caption{First experiment with ground and aerial robots for the
    PDVRP problem.
    The red pins represent pickups, while the blue ones represent deliveries.
    The dashed lines represent the paths to follow.}
  \label{fig:pdvrp_experiment_1}
\end{figure}

In the next, we report the results of an experiment for the pickup-and-delivery problem solved using the Distributed Primal Decomposition scheme summarized in Algorithm~\ref{alg:algorithm_primal_milp}~\cite{camisa2022multi}. Here, we employed a real team of heterogeneous robots including both Crazyflie nano-quadrotors and Turtlebot3 mobile robots.
We consider an experimental setting in which a set of $10$ pickup and $10$ delivery tasks must be served by the team of robots. A task is considered accomplished once the designated robot reaches the task position. To simulate the load/unload of goods in classical pickup-and-delivery settings, robots have to wait on the task for a certain random time, which is drawn uniformly between $3$ and $5$ seconds. Similarly, the
capacity of each robot and the demand/supply of the requests
are drawn from uniform distributions. Also, each robot can serve only a subset of the overall number of tasks.
The designated robotic fleet is composed of $2$ Crazyflie nano-quadrotors
and $7$ TurtleBot3 Burger mobile robots. At the beginning of the experiment, robots are randomly placed in the environment. As for the low-level control routines, we follow the setting discussed in Section~\ref{sec:exp_cco1} 
to avoid collisions among mobile, ground robots. As for the quadrotors instead, we implemented a flatness-based hierarchical control scheme.
A snapshot from the experiment is provided in Figure~\ref{fig:pdvrp_experiment_1}
\footnote{A video can be found at \texttt{\url{https://youtu.be/NwqzIEBNIS4}}.}.

\subsection{Experiments for \aggropt{1}}
\label{sec:aggro1_exp}

\begin{figure}[htbp]
  \centering
  \includegraphics[width=0.49\columnwidth]{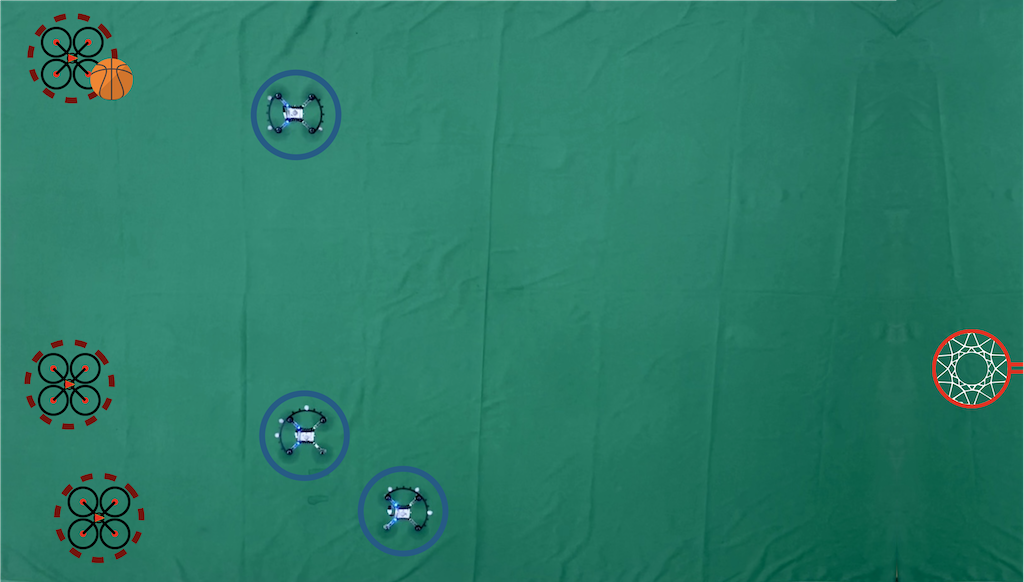}
  \includegraphics[width=0.49\columnwidth]{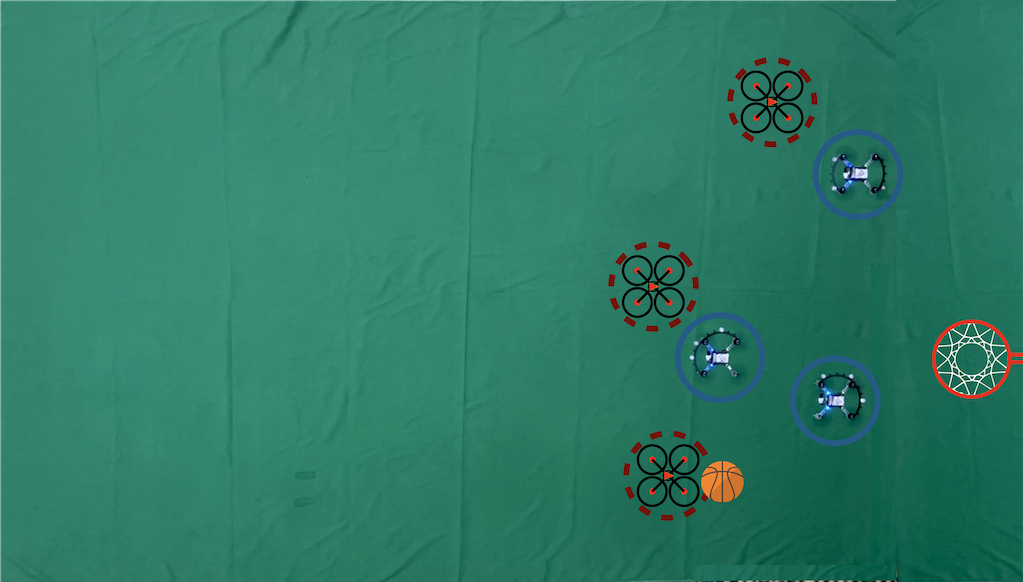}
  \caption{Snapshot from an experiment. Real defenders are highlighted by blue circles and
    virtual attackers are depicted in red.
  }
  \label{fig:snapshot_exp}
\end{figure}

Here, we report the experiments provided in~\cite{pichierri2023distributed}.
Specifically, we recast the surveillance framework detailed in Section~\ref{sec:aggro1} to provide experiments for a multi-robot basketball match.  Here, both real robots and virtual Crazyflie nano-quadrotors are used. The virtual ones are simulated leveraging the Webots environment. We point out that robots in the Webots environment are simulated up to the motor level, and firmware bindings are used to perform a realistic simulation.
The scenario unfolds as follows. A fleet of $\N = 3$ robots plays the role of the defending team, with the defensive strategy modeled as an aggregative optimization problem. The offending team,  also composed of $3$ robots, moves following pre-defined trajectories. 
In this setting, the optimization variable $\xag[i][t] \in \real^3$ models the position of robot $i\in\set$ at time $t$. Each robot $i\in\set$ has to man-mark a robot in the offending team, whose position is denoted as $p_{i}^t \in \real^3$. The position of the ball, known by all the robots, is denoted by $b^t \in \real^3$. Similarly, the position of the basket is denoted by $p_{\text{bsk}} \in \real^3$.
The generic robot $i$ in the defending team aims at placing over the segment linking $p_{i}^t$ and $p_{\text{bsk}}$. Also, robots in the defending team want to steer their barycenter on the segment linking the basket and the ball. This strategy can be cast as an aggregative optimization problem in the form of~\eqref{eq:aggregative_problem}, with objective functions
\begin{align}
  \fag[i][t](\xag[i],\sigma(\xag),\xjneigh)&= \gamma_{i,p}\norm{x_i - \tilde{p}_i^t}^2 + \gamma_{\text{agg}}\norm{\sigma - \tilde{b}^t}^2 
                                             \notag\\
                                           &\hspace{.5cm}
                                             + \sum_{j\in \innbrs_i^t} -\log(\Vert\xag[i]-\xag[j]\Vert),
                                             \label{eq:basket}
\end{align}
where $\gamma_{i,p}, \gamma_{\text{agg}} \in [0,1]$. Also, $\tilde{p}_i^t = \lambda_i p_{\text{bsk}} + (1-\lambda_i)p_i^t$ models a point on the segment connecting the basket and the $i$-th offensive player. Similarly,  $\tilde{b}^t = (1-\lambda_{\text{agg}})p_{\text{bsk}} + \lambda_{\text{agg}}b^t$ models a point on the segment connecting the ball to the basket. 
It is worth noting that, differently from the setup in Section~\ref{sec:aggregative}, the cost function~\eqref{eq:basket} has also a term $\sum_{j\in \innbrs_i^t} -\log(\Vert\xag[i]-\xag[j]\Vert)$ taking into account the position of other neighboring robots. This particular function models a barrier to avoid inter-robot collisions in the defending team.
In the proposed experiments, we choose as robotic platform the Crazyflie~$2.0$ nano-quadrotor. 
Intruders are virtual, simulated quadrotors. 
The described setup is tackled by implementing a distributed method, proposed in~\cite{pichierri2023distributed}, extending Algorithm~\ref{alg:pat} to deal with (i) time-varying networks and (ii) the barrier functions~\eqref{eq:basket}.
Moreover, in order to predict the updated positions of the target and the intruders, a Kalman-based mechanism is interlaced with the optimization scheme.
A snapshot from an experiment is in Figure~\ref{fig:snapshot_exp}
\footnote{The video is available at \url{https://www.youtube.com/watch?v=5bFFdURhTYs}.}.

\subsection{Simulations for \aggropt{2}}

\begin{figure}
  \includegraphics[width=\columnwidth]{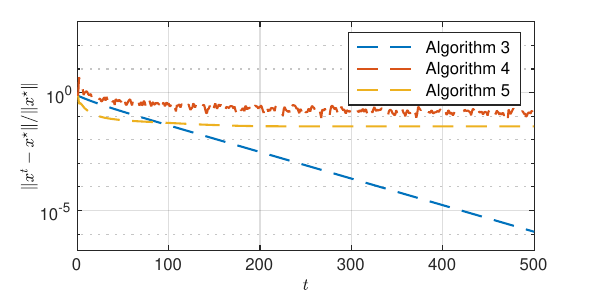}
  \caption{Relative optimality error convergence of Algorithms~\ref{alg:pat},~\ref{alg:FW}, and~\ref{alg:ADMM_aggr}.}
  \label{fig:aggro2}
\end{figure}

\begin{figure}
  \includegraphics[width=\columnwidth]{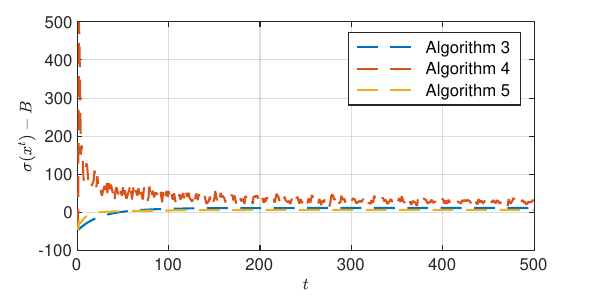}
  \caption{Evolution of $\sigma(\xag[][t]) - B$ in Algorithms~\ref{alg:pat},~\ref{alg:FW}, and~5.}
  \label{fig:aggro2_cc}
\end{figure}

In this section, we provide numerical simulations addressing the multi-robot resource allocation scenario described in Section~\ref{sec:aggro2}, i.e., the one formulated as an instance of the aggregative optimization problem as given in~\eqref{eq:aggregative_problem}.
To tackle this problem in a distributed fashion, we implement Algorithm~\ref{alg:pat}, Algorithm~\ref{alg:FW}, and Algorithm~\ref{alg:ADMM_aggr} (cf. Section~\ref{sec:aggregative_algo}).
The numerical simulation is organized as follows.
A team of $N = 10$ robots cooperatively allocates a common, scalar resource $\sigma(\xag) = \sum_{i=1}^\N \xag[i]$ according to a common performance criterion. 
Specifically, the robots have to perform a set of individual tasks and, for each robot $i \in \until{10}$, there is a quadratic utility function $U_i: \real \to \real$ assessing the quality of of the service provided by robot $i$ according to
\begin{align}
  U_i(\xag[i]) = \frac{1}{2}q_i^2\xag[i][2] + \xag[i]r_i,
\end{align}
where $q_i > 0$ and $r_i \in \real$ for all $i \in \until{10}$.
Further, given $B > 0$, the penalty function $p: \real \to \real$ is defined as
\begin{align*}
  p(\sigma(x) - B) = \exp(\sigma(x) - B).
\end{align*}
Therefore, for all $i \in \until{10}$, the local objective function $\fag[i]$ reads as
\begin{align*}
  f_i(\xag[i],\sigma(\xag)) =  \frac{1}{2}q_i^2\xag[i][2] + \xag[i]r_i + \frac{\exp(\sigma(x) - B)}{\N},
\end{align*}
Moreover, each quantity $\xag[i]$ is constrained to belong to $X_i \triangleq \{\xag[i] \in \real \mid 0\leq \xag[i]\leq x^{\text{max}}\}$ for some $x^{\text{max}} > 0$.
For all $i \in \until{10}$, the parameters $q_i$ and $r_i$ are randomly generated extracting them, with uniform probability, from the intervals $[1,10]$ and $[-100,0]$, respectively.
Then, we set $B = 100$ and $x^{\text{max}} = 100$.
As for the communication among the robots of the network, we randomly generate an undirected, connected \er/ graph with parameter $0.1$ and an associated doubly stochastic weighted adjacency matrix.
As for the algorithms' parameters, we implement (i) Algorithm~\ref{alg:pat} choosing $\delta = \gamma = 0.01$, (ii) Algorithm~\ref{alg:FW} choosing $\gamma^t = 1/\sqrt{t}$, and (iii) Algorithm~\ref{alg:ADMM_aggr} choosing $\xi = \rho =0.1$.
Figure~\ref{fig:aggro2} reports the performance of the algorithms in terms of the convergence of the relative optimality error $\frac{\norm{\xag[][t] - x^\star}}{\norm{\xag[][\star]}}$, where $\xag[][\star] \in \real^{10}$ is the (unique) solution to the considered problem.
As predicted by Theorem~\ref{th:pat}, Theorem~\ref{th:FW}, and~\ref{th:grontas}, Figure~\ref{fig:aggro2} shows that the three algorithms converge toward $\xag[][\star]$. 
However, coherently with the claims of Theorem~\ref{th:pat}, Theorem~\ref{th:FW}, and~\ref{th:grontas}, the linear convergence is only achieved by Algorithm~\ref{alg:pat}.
Finally, in Figure~\ref{fig:aggro2_cc}, we report the evolution over $t$ of the quantity $\sigma(\xag[][t]) - B$ achieved by running Algorithm~\ref{alg:pat}, Algorithm~\ref{alg:FW}, and~\ref{alg:ADMM_aggr}.
As one may expect, such a quantity is not necessarily negative because the exceeding configurations are not forbidden but only penalized.

  \section{Research Directions}
  \label{sec:research_dir}
  
  In this section, we present some research directions extending the scenarios considered in this
  tutorial.

  \paragraph{Nonconvex optimization problems}
  \label{sec:nonconvex}
  
  In some multi-robot tasks, the convexity properties required in Assumption~\ref{ass:dualdec_convex},~\ref{ass:convexity},~\ref{ass:objective_FW}, and~\ref{ass:proper_func} may be not realistic.
  Avoiding these kinds of assumptions significantly complicates the algorithms' analysis and weakens the theoretical guarantees since, typically, one can only ensure convergence towards the set of problem stationary points, thus including also local minima.
  Only a few works consider nonconvex settings in the distributed optimization community, see, e.g.,~\cite{daneshmand2020second,carnevale2022nonconvex} for consensus optimization problems,~\cite{camisa2021distributed_b,camisa2019distributed} in the constraint-coupled scenario, and~\cite{carnevale2023nonconvex} for the aggregative one.
  
  \paragraph{Mixed-integer and discrete optimization problems}
Several tasks in robotic networks, such as trajectory planning with obstacle avoidance, task allocation, and sensor placement that can be cast as combinatorial optimization problems in which the decision variables (or a subset of them) may belong to discrete sets. Two well-known classes of problems, namely Mixed-Integer Linear Problems and Submodular Problems, have started to receive attention from the Robotics community. There are special classes of MILPs in which it is possible to leverage convex optimization techniques to solve these problems. An example is the assignment problem considered in Section~\ref{sec:task_allocation} that can be solved as a linear program~\cite{burger2012distributed}. Note that this property is in general not valid. Indeed, neglecting the integer constraints while solving a (mixed-)integer problem yields to solutions that may not be integer-valued. An example of a problem for which this property does not hold is the PDVRP in Section~\ref{sec:pdvrp}.
In general, finding an optimal solution to MILPs is computationally hard. To this end, several strategies resort to sub-optimal schemes with fast convergence properties. Distributed algorithms for mixed-integer problems with provable sub-optimality bounds have been proposed in~\cite{camisa2021distributed,falsone2018distributed}. A related approach without optimality guarantees has been proposed in~\cite{manieri2023novel}. Authors in~\cite{testa2021generalized} propose a distributed branch-and-price scheme for the Generalized Assignment Problem. The procedure allows us to find an optimal solution to the problem or to early stop and retrieve a sub-optimal solution. A class of MILPs in which robots share a common optimization variable but have local private constraints is considered in~\cite{manieri2021hyper,testa2019distributed,testa2017finite}. This setup is different from the ones considered in this paper, and ad-hoc methodologies are employed to solve the optimization problem. As for Submodular Optimization, the idea is to minimize a set-based function. Thus, the optimization variable is a set itself. For this reason, there is a need to develop ad-hoc distributed methodologies to solve these problems. Submodular minimization can be solved exactly by leveraging convex optimization techniques. However, it has received few attention in the distributed optimization literature~\cite{testa2020distributed,jaleel2019distributed,jaleel2018real,testa2018distributed}. On the other hand, submodular maximization is typically solved using sub-optimal, greedy algorithms with performance guarantees. This problem has been investigated more in detail by the distributed optimization community, we refer the reader to~\cite{zhou2022distributed,grimsman2018impact,gharesifard2017distributed,williams2017decentralized} for recent references. In these setups, network agents converge to a common optimal set. Thus, the methods discussed in this paper cannot be easily applied.
  
  \paragraph{Imperfect communication protocols} 
  In several applications involving networks of robots, assuming a perfectly synchronous and
  reliable communication protocol among robots may be not realistic.
  For this reason, it would be useful to either design reliable communication protocols or design algorithms that are robust with respect to asynchronous exchanges and/or packet losses in the communication among robots.
  In an asynchronous setting, each robot $i \in \set$ updates and/or sends its
  local quantities in a totally uncoordinated manner
   with respect to the other
  robots of the network.
  In the distributed optimization literature, such a setting has already gained attention from a theoretical point of view but mainly in the consensus optimization setup, see, e.g.,~\cite{notarnicola2016asynchronous,notarnicola2016randomized,notarnicola2017distributed,tian2018asy,zhang2019asyspa,tian2020achieving,bastianello2020asynchronous,jiang2021asynchronous,carnevale2023triggered}.
  As for networks with packet losses, they are often modeled by considering that a message sent by robot $i$ is successfully received by robot $j$ with a given probability.
  Preliminary theoretical results dealing with networks subject to packet losses are available in~\cite{bof2018multiagent,lei2018asymptotic,bastianello2018distributed,bastianello2020asynchronous}.
  Several challenges are still open at a theoretical level for the presented (constraint-coupled and aggregative) optimization frameworks. Moreover, handling realistic communication models in specific robotic applications is mainly an open problem.

  \paragraph{Unknown functions and personalized optimization scenarios} 
  There has been in the last years an increasing interest in scenarios in which the objective function is completely or partially unknown.
  These scenarios arise, e.g., when robots take measurements via onboard sensors and the function profile is not explicitly available. 
  When functions are not known, gradient-based algorithms are not implementable. To overcome this issue, extremum-seeking and zero-order techniques have been proposed to solve the optimization problem. Several applications have been considered, including, e.g., source seeking~\cite{li2020cooperative,khong2014multi} and resource allocation~\cite{poveda2013distributed,wang2019distributed}, object manipulation~\cite{calli2018active}, and trajectory planning~\cite{bagheri2018multivariable,jain2021optimal}. These algorithms have been studied from a theoretical perspective in the distributed framework,~\cite{menon2014collaborative,guay2021distributed,mimmo2021extremum,tang2020distributed,liu2014sample,yuan2014randomized}. However, there are few applications of these distributed methodologies in multi-robot settings.
  Another interesting scenario arises when a portion of the local functions is known, while another part related to, e.g., user preferences is unknown. 
  In robotics, this may be due to the presence of humans, interacting with robots, that introduce non-engineering cost functions in the optimization problem. 
  Addressing this scenario for the constraint-coupled and aggregative optimization frameworks discussed in Sections~\ref{sec:constraint_coupled} and~\ref{sec:aggregative} is an open research direction.

  For the constraint-coupled scenario, the cost function in problem \eqref{eq:constr_coupled} becomes $\sum_{i=1}^\N (V_i(\xag[i]) + U_i(\xag[i]))$, 
  %
  %
  where $V_i :\real^{\nag[i]} \to \real$ and $U_i :\real^{\nag[i]} \to \real$ are respectively the known and unknown parts of the local objective function of robot $i$.
  %
  Analogously, for the aggregative framework, the cost function in problem \eqref{eq:aggregative_problem} becomes $\sum_{i =1}^\N (V_{i}(\xag[i],\sigma(\xag)) + U_{i}(\xag[i],\sigma(\xag)))$, with the same meaning for the two functions.
  %
  %
  %
  %
  %
  Interesting examples in robotics can be found, e.g., in the context of trajectory generation~\cite{kruse2013human,luo2020socially,zhou2021human}, rehabilitation robots~\cite{menner2020using}, and demand response tasks in energy networks~\cite{chatupromwong2012optimization,ospina2020personalized}.
  Here, human preferences are hard to model and, thus, a possible strategy to deal with them consists in resorting to the so-called \emph{personalized} approach, i.e., to progressively estimate the unknown objective function by using human feedback.
  In a centralized setting, personalized optimization has been originally addressed in~\cite{simonetto2021personalized, ospina2022time}.
  As for the distributed setup, personalized strategies are used
  in~\cite{notarnicola2022distributed} to deal with consensus optimization problems and in~\cite{carnevale2022learning} to address the aggregative scenario.
  Finally, in~\cite{fabiani2022learning}, the personalized framework is addressed in the context of game theory.

  \paragraph{Online and stochastic scenarios} Many applications arising in robotic scenarios occur in dynamic environments and, hence, need to be formalized by resorting to online and/or stochastic problems.
  Indeed, in the online setup, the cost function and constraints vary over time. That is, for the constraint-coupled setup we have  $\fag[i][t]$, $g_i^t$ and $\X[i]^t$, which are revealed to robot $i$ only once its decision $\xag[i][t]$ has been taken. Similarly, for the aggregative case $\fag[i][t]$, $\sigma^t = \sum_{i=1}^\N \phi_i^t$, and $\X[i]^t$.
  The goal is to design algorithms involving at each $t$ only one iteration step (e.g., a gradient update) without solving the entire problem revealed at $t$. 
  In this context, algorithms are commonly evaluated through the so-called \emph{dynamic regret} $R_T$.
  For the constraint-coupled framework, $R_T$ reads as 
  \begin{align*}
    &R_T \triangleq \sum_{t=1}^T \sum_{i=1}^\N(\fag[i][t](\xag[i][t]) -\fag[i][t](\xag[i,\star][t])),
  \end{align*}
  where $T \in \natural$ denotes the time horizon, while $\xag[\star][t] \triangleq \col(\xag[1,\star][t],\dots,\xag[N,\star][t])$ and $\xag[][t] \triangleq \col(\xag[1][t],\dots,\xag[N][t])$ denote respectively the minimizer of $\sum_{i=1}^\N \fag[i][t]$ and the corresponding robots' estimate for all $t \in \{1,\dots,T\}$.   
  Analogously, for the aggregative setting, $R_T$ is defined as 
  \begin{align*}
    R_T \triangleq \sum_{t=1}^T \sum_{i=1}^\N(\fag[i][t](\xag[i][t],\sigma^t(\xag[][t])) - \fag[i][t](\xag[i,\star][t],\sigma^t(\xag[\star][t]))).
  \end{align*}
  %
  %
  Distributed algorithms for online constraint-coupled optimization can be found in \cite{gu2019adaptive,yi2019distributed,yi2020distributed,li2020online,li2020distributedconstraint_a,li2020distributedconstraint_b}, while
  %
  %
  for online aggregative problems can be found in \cite{li2021distributed_b,carnevale2021distributed,carnevale2022learning,pichierri2023distributed}.
  In the stochastic versions of the above problems, the variation over $t$ of these functions and sets is due to the realization at iteration $t$ of a stochastic process, see, e.g.,\cite{camisa2021stochastic} for the constraint-coupled framework or \cite{li2021distributed_b} for the aggregative one.

  Recently, in the context of online and/or stochastic optimization, novel methods embedding prediction schemes have gained attention for centralized settings~\cite{simonetto2016class,simonetto2017prediction,simonetto2018dual,bastianello2022opreg} or with consensus schemes~\cite{cao2024multi}. Designing distributed optimization algorithms for the above frameworks is an interesting research direction. Moreover, implementing enhanced schemes in multi-robot scenarios is mainly an open problem.

\section{Conclusion}

In this paper, we considered cooperative robotics from the point of view of distributed optimization.
Specifically, we focused on constraint-coupled and aggregative optimization setups and showed how to recast several challenging problems arising in multi-robot applications into these distributed optimization settings.
Then, for both these setups, we reviewed different distributed algorithms to solve them and provided their convergence properties.
Moreover, we revised toolboxes to implement distributed optimization schemes on real, multi-robot networks. 
To bridge the theoretical analysis and the toolbox presentation, we provided simulations and experiments on real teams of heterogeneous robots for different use cases.


\end{document}